\documentclass{article}

\usepackage[final]{corl_2026}% Use this for the initial submission.
\usepackage{booktabs}  % \toprule, \midrule, \bottomrule, \cmidrule
\usepackage{multirow}  % \multirow
\usepackage{xspace}
\usepackage{graphicx}
\usepackage{amsmath}
\usepackage{amssymb}
\usepackage{wrapfig}
\usepackage{adjustbox}
\usepackage{titlesec}

\definecolor{myGreen}{rgb}{0.02,0.5,0.02}
\definecolor{myRed}{rgb}{0.8, .2, .2}
\definecolor{myBlue}{rgb}{0.0,0.22,0.8}
\definecolor{myBlack}{rgb}{0.0, 0.0, 0.0}

\title{FreeLoc: Online Floorplan Localization via Diffusion-Aided Pose Refinement}

\author{
\textbf{Haocheng Peng,
Boyang Zhou,
Jiarui Hu$^{\dagger}$,
Xiyue Guo,
Ziyang Zhang,}\\
\textbf{Boming Zhao,
Yifan Gao,
Xiao Li,
Hujun Bao,
Zhaopeng Cui$^{\ddagger}$}\\[1mm]
State Key Lab of CAD\&CG, Zhejiang University\\[3mm]
\textbf{Project Website: }
\href{https://zju3dv.github.io/freeloc/}
{\textcolor{blue}{\texttt{zju3dv.github.io/freeloc/}}}
}
\begin{document}
\maketitle

\begingroup
\renewcommand{\thefootnote}{\fnsymbol{footnote}}
\footnotetext[2]{Project leader}
\footnotetext[3]{Corresponding author}
\endgroup

\newcommand{\ours}{FreeLoc\xspace}
\newcommand{\fthreeloc}{F$^3$Loc}
\newcommand{\green}[1]{\textcolor{green}{#1}}

%control section spacing
% \newcommand{\compactsectionspacing}{
%     \titlespacing*{\section}{0pt}{0.65em}{0.25em}
%     \titlespacing*{\subsection}{0pt}{0.45em}{0.20em}
%     \titlespacing*{\subsubsection}{0pt}{0.35em}{0.15em}
% }

%===============================================================================

\begin{abstract}
     Floorplans provide compact and widely available geometric maps for indoor localization, but existing high-performing floorplan-based methods still convert them into dense scene-specific offline databases, tying accuracy, storage, and runtime to the sampling resolution of the discretized pose space. We present \ours, an online RGB-based floorplan localization framework that treats the floorplan as a directly queryable geometric map. \ours introduces an efficient online geometric querying and diffusion-aided refinement scheme, which retrieves plausible pose anchors through on-the-fly floorplan ray querying and refines them into accurate continuous pose estimates. For sequential localization, \ours develops an online likelihood construction strategy that bridges single-frame localization and probabilistic temporal fusion by constructing likelihoods from coarse-sampled candidates and refined pose hypotheses, enabling histogram-filter-based temporal fusion without offline databases. Experiments demonstrate real-time online inference and state-of-the-art performance in both single-frame and sequential localization, while real-world results validate practical deployability in indoor robotic localization scenarios.

\end{abstract}

\keywords{Floorplan Localization, Indoor Localization, Pose Refinement} 

\begin{figure*}[!htbp]
    \centering
    \vspace{-0.55em}
    \includegraphics[width=0.85\textwidth]{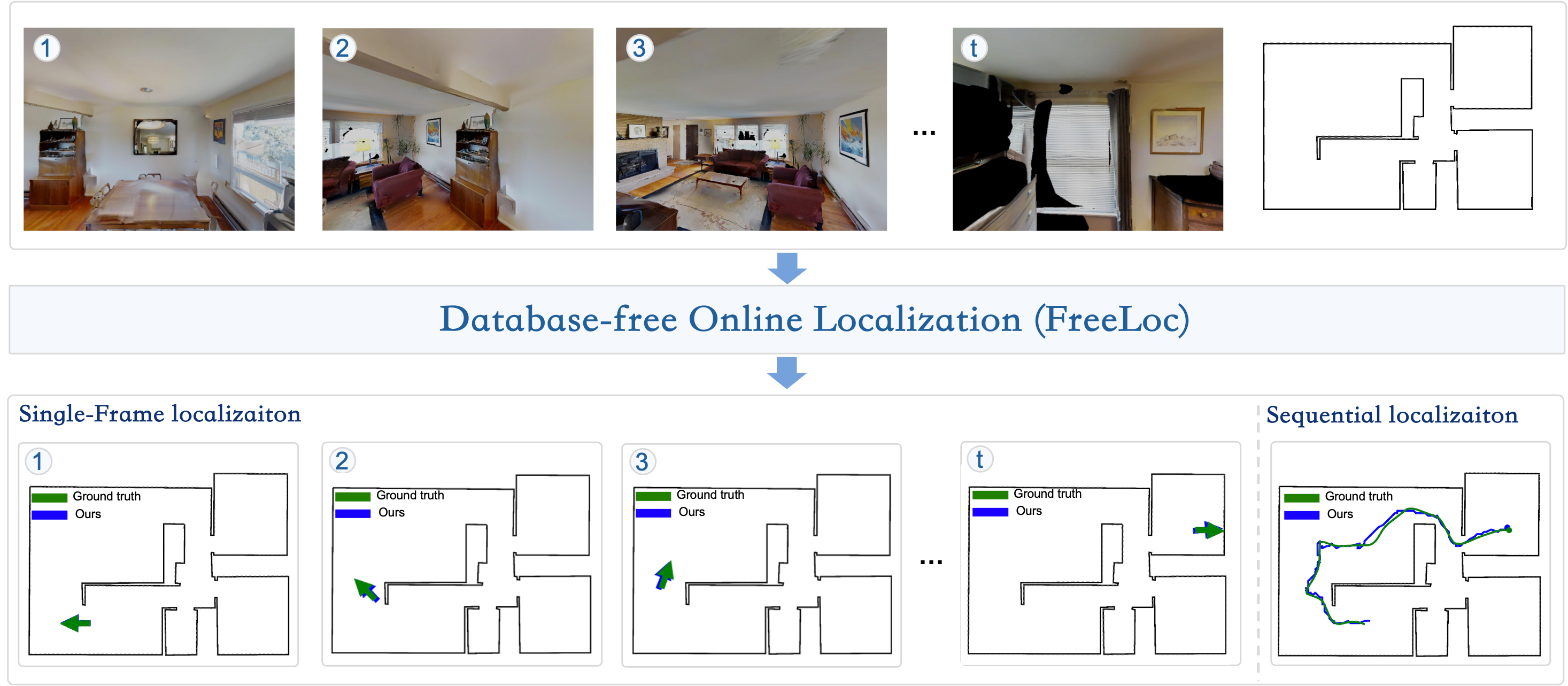}
    \vspace{-0.5em}
    \caption{\ours enables database-free online floorplan localization, directly using the 
provided floorplan to support accurate single-frame and 
sequential localization in new environments.}
    \label{fig:teaser}
    \vspace{-0.5em}
\end{figure*}
%===============================================================================

%\compactsectionspacing

\section{Introduction}
	
    Indoor camera localization is a core capability for robotic perception and embodied intelligence, providing the spatial grounding required for mobile robot navigation, human-robot interaction, and autonomous exploration. For robots operating in real indoor environments, a localization system should not only be accurate, but also be readily deployable in new scenes with minimal scene-specific preparation. Traditional indoor visual localization methods~\cite{img_retrieval:kalantidis2011viral,img_retrieval:berton2025megaloc,3dmodel-based:liu2017efficient,3dmodel-based:sarlin2019coarse} typically rely on pre-built image databases or accurate 3D scene models. These representations are costly to construct, storage-intensive, and difficult to maintain, making them difficult to scale across many buildings and adapt to changing indoor environments.

Floorplans offer a lightweight and scalable map representation for indoor localization. Compared with image databases and 3D scene models, they are compact, widely available~\cite{floorplan_recon:Gueze_2023_ICCV, otten1982automatic}, and relatively stable over time, making them attractive for localization systems deployed across many buildings. Ideally, a floorplan should serve as a directly queryable geometric map, allowing localization without constructing additional scene-specific databases. However, existing high-performing floorplan localization methods~\cite{chen2024f3loc, semantic-based:grader2025supercharging, wuest2025unloc} still largely rely on a database-driven formulation: they discretize the $\mathrm{SE}(2)$ pose space, precompute dense pose-indexed observations from the floorplan, and match query images against these entries for localization. In sequential settings, the same database-derived scores are further converted into observation likelihoods for histogram filtering, causing the database dependency to persist in temporal fusion. As a result, the limitation lies not merely in preprocessing cost, but in the discretized localization paradigm itself: once localization is tied to a finite pose-indexed database, localization precision, storage, and runtime all scale with the sampling resolution, creating an unfavorable trade-off among accuracy, efficiency, and scalability.

This motivates a more practical and challenging setting: \emph{given only a floorplan, can we directly query it online to achieve accurate RGB-based single-frame and sequential localization in new environments, without constructing a scene-specific offline database?}

A useful clue comes from how humans localize themselves in unfamiliar indoor spaces. Rather than exhaustively comparing the current view with every possible location on the map, humans typically use the floorplan to narrow the search to a few plausible regions, and then refine their location by matching local geometric details with the current observation. However, realizing this coarse-to-fine intuition in an online localization system is non-trivial. 
First, the system must rapidly search
the entire floorplan pose space online to identify plausible initial pose
hypotheses. Second, these retrieved poses are inherently approximate: finite
online sampling introduces discretization errors, while noisy and incomplete
RGB-derived geometric observations further limit matching accuracy, making it
challenging to recover accurate poses from such coarse hypotheses. Third, for
sequential localization with histogram filtering, observation likelihoods over
the discretized pose space are required to update the propagated motion prior,
yet these likelihoods can no longer be densely evaluated from a precomputed
scene-specific ray database.
%Online floorplan querying must avoid exhaustive dense sampling while maintaining sufficient coverage for ambiguous indoor layouts. Meanwhile, geometric cues inferred from a single RGB image are often noisy and partial due to occlusions and limited field of view, while coarse online sampling inevitably introduces discretization errors. As a result, the retrieved poses are usually approximate and may still deviate 
%from the true camera pose. 
%For sequential localization, the challenge further extends to constructing observation likelihoods over the pose space, since these likelihoods can no longer be densely evaluated from a precomputed scene-specific ray database.
%For sequential localization, the challenge further extends to probabilistic temporal fusion: Bayesian filtering requires observation likelihoods over the pose space to update the propagated motion prior, but these likelihoods can no longer be densely evaluated from a precomputed scene-specific ray database.

To address these challenges, we propose FreeLoc, an online RGB-based floorplan localization framework that treats the floorplan as a directly queryable geometric map rather than a scene-specific offline retrieval database, as shown in Fig.~\ref{fig:teaser}. FreeLoc first retrieves a compact set of plausible pose anchors by querying geometric rays from the floorplan on the fly using a GPU-accelerated rotation-aware ray kernel. To refine these approximate anchors, we introduce a diffusion-aided pose refinement network that estimates residual pose corrections from the discrepancy between image-side and floorplan-side ray observations. Instead of relying on one-shot deterministic regression, we formulate pose 
refinement as an iterative denoising process, which better matches the nature 
of residual pose optimization and progressively corrects imperfect coarse 
anchors under ray-level geometric guidance. This allows FreeLoc to move beyond the resolution of coarse online samples and produce accurate pose hypotheses in continuous pose space without exhaustive dense sampling or precomputed scene-specific ray databases. To support sequential localization, we further integrate the online coarse-to-fine estimator into a probabilistic filtering pipeline. Rather than deriving dense observation likelihoods from an offline ray database, FreeLoc constructs observation likelihoods online from coarse-sampled candidates and refines their high-confidence regions using the refined pose hypotheses before Bayesian filtering~\cite{chen2003bayesian, fox2003bayesian,dellaert1999monte}. This online likelihood construction bridges single-frame localization and temporal fusion without reintroducing a scene-specific database, enabling temporally consistent localization while preserving the immediate deployability of floorplans.

Our contributions are summarized as follows. First, we propose FreeLoc, an online coarse-to-fine framework for RGB-based floorplan localization that eliminates scene-specific offline database construction and enables immediate deployment in new environments. Second, we introduce an efficient online geometric querying and diffusion-aided refinement scheme, which retrieves plausible pose anchors directly from the floorplan and progressively refines them into accurate continuous pose estimates. 
%Third, we develop an online likelihood construction strategy that bridges single-frame localization and probabilistic temporal fusion, enabling histogram-filter-based sequential localization by constructing observation likelihoods from online sampled candidates and updating them with refined pose hypotheses.
Third, we develop an online likelihood construction strategy that bridges single-frame localization and probabilistic temporal fusion, enabling histogram-filter-based sequential localization by constructing observation likelihoods from online sampled candidates and updating them with refined pose hypotheses.
Finally, experiments on various datasets demonstrate that FreeLoc achieves SOTA accuracy and efficiency across representative localization methods from different paradigms, while real-world experiments further validate its practicality for indoor robotic localization.

%===============================================================================

\section{Related Work}
	
    \textbf{Visual Localization.} 
%Visual localization is a fundamental problem in robotic perception and has been extensively studied over the past decades. Early approaches typically formulate localization as an image retrieval problem, where a query image is matched against a pre-built reference database and its pose is inferred from the most similar retrieved images. 
%With the development of 3D vision, another line of work constructs explicit 3D scene models, often through structure-from-motion (SfM), and estimates the 6-DoF camera pose by establishing 2D--3D correspondences between the query image and the reconstructed model. 
%Different from correspondence-based pipelines, scene coordinate regression methods predict the 3D scene coordinate for each image pixel, while pose regression methods directly regress the camera pose using neural networks. 
%More recently, with the emergence of neural scene representations such as NeRF and 3D Gaussian Splatting, increasing efforts have explored using these representations to relate query images to the underlying scene geometry and recover camera poses. 
%However,  these approaches depend on pre-constructed databases or scene-specific 3D models, leading to high storage costs and limited generalization to unseen environments.
Visual localization is a fundamental problem in robotic perception and has 
been extensively studied for decades. 
Early methods commonly 
formulate it as 
image retrieval~\cite{img_retrieval:kalantidis2011viral, img_retrieval:berton2025megaloc,img_retrieval:arandjelovic2016netvlad,img_retrieval:chum2007total, img_retrieval:keetha2023anyloc}, where a query image is matched to a pre-built reference 
database to infer its pose. With advances in 3D vision, correspondence-based 
approaches~\cite{3dmodel-based:liu2017efficient,3dmodel-based:sarlin2019coarse,3dmodel-based:kim2023ep2p,3dmodel-based:Lindenberger_2021_ICCV} construct %explicit 
3D scene models, typically via 
structure-from-motion (SfM)~\cite{ozyecsil2017survey}, and estimate the 6-DoF camera pose from 2D-3D 
matches. 
Beyond ground-level reference maps, cross-view localization
has also explored globally available satellite imagery as a spatial reference
for ground-level localization~\cite{guo2024satellite}.
%Other learning-based methods regress scene coordinates or camera poses directly from images. 
More recently, neural scene representations such as NeRF~\cite{mildenhall2021nerf} 
and 3DGS~\cite{3dgaussian} have also been explored for pose estimation by 
relating query images to scene-specific representations~\cite{nerf-guassian-based:zhao2024pnerfloc,nerf-guassian-based:zhai2025splatloc,nerf-guassian-based:huang2025sparse}. However,  these approaches depend on pre-constructed databases or scene-specific representations, leading to high storage costs and limited generalization to unseen environments.

\textbf{Floorplan Localization.} Floorplan localization estimates camera pose with respect to a given floorplan, offering a lightweight alternative to image databases and full 3D reconstructions. 
Early methods~\cite{fp_Lidar-based:boniardi2017robust, fp_Lidar-based:wang2019glfp,fp_Lidar-based:FP-Loc,fp_Lidar-based:10321673} rely on LiDAR scans matched against floorplan geometry, while later approaches~\cite{rgbd_vo_based:ito2014w,rgbd_vo_based:Chu_2015_ICCV,rgbd_vo_based:Wijmans_2017_CVPR,rgbd_vo_based:winterhalter2015accurate} use depth cameras or visual odometry to obtain geometric observations. 
Although effective, these methods require dedicated sensors or assumptions such as known room or camera height, limiting practical applicability. 
To improve accessibility, recent works have explored RGB-only floorplan localization. 
LaLaLoc~\cite{howard2021lalaloc} and LaLaLoc++~\cite{howard2022lalaloc++} embed panoramic observations and floorplan-derived layouts into a shared latent space for retrieval. 
PF-Net~\cite{pf-net} formulates localization as differentiable particle filtering, and LASER~\cite{semantic-based:min2022laser} aligns image features with rotation-aware circular map features rendered from rasterized floorplan geometry. 
However, many of these methods assume upright camera poses, making them less suitable for handheld or head-mounted settings. 
Another line of work~\cite{semantic-based:grader2025supercharging,semantic-based:min2022laser,semantic-based:mendez2020sedar,semantic-based:zimmerman2022long} incorporates semantic cues, but semantic annotations are often unavailable in standard floorplans. 
\fthreeloc~\cite{chen2024f3loc} and UnLoc~\cite{wuest2025unloc} take an important step toward practical RGB-based floorplan localization by using geometry-only floorplans and images captured under arbitrary camera poses. 
They predict floorplan depth from gravity-aligned images and localize the camera by matching the resulting rays against a dense pre-sampled floorplan-ray database. 
However, constructing such a database requires time-consuming scene-specific preprocessing and creates an inherent trade-off among localization accuracy, sampling density, and storage cost, limiting flexibility in new environments. 
In contrast, our method performs online localization without scene-specific offline ray databases, enabling efficient and accurate deployment in new scenes using only the provided geometric floorplan.

\section{Method}
\label{sec:method}
\begin{figure*}[t]
    \centering
    \includegraphics[width=1\textwidth]{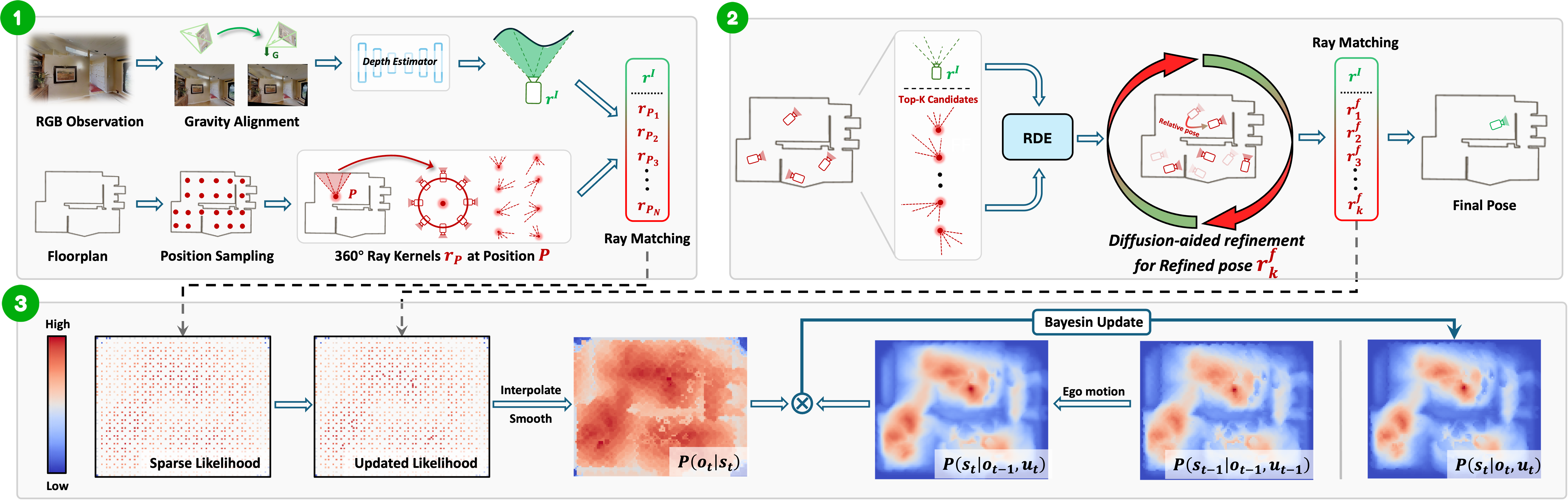}
    \vspace{-2em}
    \caption{\textbf{Overview of \ours.} 
    %We retrieve top-$K$ coarse anchors by matching scan rays from the gravity-aligned RGB image with online queried floorplan rays. A diffusion-aided module refines these anchors, and the refined hypotheses are re-scored for the final single-frame estimate. For sequential localization, online likelihoods are constructed from coarse and refined candidates and fused with a histogram filter for temporally consistent pose estimation.
We retrieve top-$K$ coarse anchors by matching image-side rays with online queried floorplan rays, then refine and re-score them for the final estimate.
For sequential localization, coarse and refined candidates construct online likelihoods for 
%histogram-filter-based 
temporal fusion.}
    % \ours performs database-free online floorplan localization with a coarse-to-fine design, supporting both single-frame pose estimation and histogram-filter-based sequential localization without scene-specific offline ray databases.our pipeline follows a coarse-to-fine design. We first convert the gravity-aligned RGB image into image-side scan rays and retrieve top-$K$ coarse pose anchors by matching them with floorplan-side rays queried online from sampled pose hypotheses. These anchors are then optimized by a diffusion-aided pose refinement module using ray-level geometric cues, producing refined hypotheses that are re-scored to obtain the final single-frame estimate, as described in Sec.~\ref{sec:single_frame_loc}. For sequential localization, we build online observation likelihoods from coarse candidates and refined hypotheses, and fuse them with a histogram filter for temporally consistent pose estimation, as introduced in Sec.~\ref{sec:seq_loc}.
    
%Given an RGB image or sequence and afloorplan, \ours performs database-free online floorplan localization with a coarse-to-fine design. It first retrieves plausible pose anchors directly from the floorplan via online coarse pose retrieval, and then refines them with diffusion-aided pose optimization. The refined hypotheses are used for single-frame localization and further converted into online observation likelihoods for histogram-filter-based sequential localization. This enables accurate single-frame and temporally consistent sequential localization without scene-specific offline ray databases.
    
    \label{fig:overview}
    \vspace{-1.0em}
\end{figure*}

In this work, we address RGB-based camera pose estimation in a known 2D floorplan.
The key idea of \ours is to treat the floorplan as an online geometric query interface rather than a precomputed pose-indexed database.
Given a floorplan map, camera intrinsics $\mathbf{K}$, gravity direction, and either a single RGB image or an RGB sequence 
$\mathcal{I}=\{I_t\}_{t=0}^{T}$ with inter-frame relative poses, our goal is to estimate the camera pose 
$\mathbf{s}_t=[s_{x,t},\,s_{y,t},\,s_{\phi,t}]\in\mathrm{SE}(2)$ 
in the floorplan coordinate system, where $(s_{x,t},s_{y,t})$ denotes the 2D camera position and 
$s_{\phi,t}$ denotes the yaw angle.
As summarized in Fig.~\ref{fig:overview}, \ours builds a unified online coarse-to-fine framework: it retrieves pose anchors through on-the-fly floorplan ray querying, refines them with diffusion-aided residual optimization, and reuses the resulting hypotheses to construct online observation likelihoods.
This enables both single-frame and sequential localization without scene-specific offline ray databases.

\subsection{Single Frame Localization}
\label{sec:single_frame_loc}

% \textbf{Online Coarse Pose Retrieval.} As the first step of our coarse-to-fine pipeline, the coarse stage performs 
% efficient database-free online pose retrieval by querying geometric 
% ray representations directly from the floorplan.
% Since query images in real-world scenarios are often captured by handheld 
% devices with arbitrary orientations, we first normalize each image using 
% gravity alignment. Given the camera intrinsic matrix $\mathbf{K}$ and the 
% roll $r$ and pitch $p$ of the image, the rotation from the camera frame to the 
% gravity-aligned frame is defined as $\mathbf{R}_{gc}=(\mathbf{R}_{z}(r)\mathbf{R}_{x}(p))^\top$.
% Note that roll and pitch are readily available from inertial sensors on modern mobile 
% devices, or can be estimated using off-the-shelf calibration methods such as 
% GeoCalib~\cite{veicht2024geocalib} when sensor readings are unavailable. 
% The image is then rectified by the homography 
% $\mathbf{H}=\mathbf{K}\mathbf{R}_{gc}\mathbf{K}^{-1}$, 
% aligning its vertical direction with gravity around the optical center. 
% The gravity-aligned image, together with a visibility mask for invalid 
% rectified regions, is used for floorplan depth estimation.

\textbf{Online Coarse Pose Retrieval.}
The first stage of \ours retrieves a compact set of plausible pose anchors by querying floorplan geometry online, instead of matching the input image against a precomputed pose-indexed ray database.
This stage provides broad pose-space coverage for ambiguous indoor layouts while keeping the candidate set sufficiently sparse for efficient refinement.

Since RGB images in practical scenarios may be captured by handheld or head-mounted devices with arbitrary roll and pitch, we first normalize each image into a gravity-aligned view.
Given the camera intrinsic matrix $\mathbf{K}$ and the image roll $r$ and pitch $p$, the rotation from the camera frame to the gravity-aligned frame is defined as 
$\mathbf{R}_{gc}=(\mathbf{R}_{z}(r)\mathbf{R}_{x}(p))^\top$.
The roll and pitch can be obtained from inertial sensors available on modern mobile devices, or estimated by off-the-shelf single-image calibration methods such as GeoCalib~\cite{veicht2024geocalib} when sensor readings are unavailable.
We then rectify the image with the homography 
$\mathbf{H}=\mathbf{K}\mathbf{R}_{gc}\mathbf{K}^{-1}$,
which aligns the image vertical direction with gravity around the optical center.
The rectified image, together with a visibility mask indicating invalid warped regions, is used to estimate floorplan depth and construct image-side ray observations.

Using the gravity-aligned image-mask pair, we follow \fthreeloc~\cite{chen2024f3loc} to obtain image-side geometric observations by predicting floorplan depth and converting it into equiangular rays $\hat{\mathbf{r}}_{t}^{I}$.
Unlike \fthreeloc~that retrieves floorplan-side rays from a precomputed pose-indexed database, we query them online directly from the geometric floorplan.
Specifically, we uniformly sample a compact set of pose hypotheses and evaluate their ray descriptors with a rotation-aware ray kernel.
For each candidate pose $(\mathbf{p},\phi)$, where $\mathbf{p}$ is the 2D position and $\phi$ is the yaw angle, the kernel casts $V_r$ equiangular rays centered at $\phi$ and computes the $i$-th ray distance by analytic boundary intersection:
%$d_i(\mathbf{p},\phi)=\min_{b\in\mathcal{B}}\rho(\mathbf{p},\phi+\Delta\theta_i,b)$,
%\vspace{-0.25em}
\begin{equation}
d_i(\mathbf{p},\phi)=
\min_{b\in\mathcal{B}}
\rho\!\left(\mathbf{p},\phi+\Delta\theta_i,b\right),
\label{eq:ray_query}
\end{equation}
where $\mathcal{B}$ denotes the set of floorplan boundary segments, $\Delta\theta_i$ is the angular offset of the $i$-th ray, and $\rho(\cdot)$ returns the nearest valid ray-segment intersection distance.

%Different from existing database-based methods~\cite{chen2024f3loc,wuest2025unloc,semantic-based:grader2025supercharging}, which rely on dense hit-testing over rasterized occupancy grids or precomputed ray databases, our query directly exploits the native line-segment structure of geometric floorplans, which substantially reduces the query cost while preserving accurate geometric ray measurements. The kernel is implemented as a vectorized batched GPU operation, allowing candidate poses and boundary segments to be evaluated in parallel. Together with compact pose sampling, this makes online floorplan ray extraction efficient enough for real-time coarse retrieval. 

Unlike existing database-based methods~\cite{chen2024f3loc,semantic-based:grader2025supercharging,wuest2025unloc}, which rely on dense hit-testing over rasterized occupancy grids, our query exploits the native line-segment structure of 
%geometric 
floorplans and computes analytic ray-boundary intersections directly. 
This %substantially 
reduces the query cost while preserving accurate geometric ray measurements.
We implement the kernel as a vectorized batched GPU operation, allowing candidate poses and boundary segments to be evaluated in parallel.
Together with compact pose sampling, this enables efficient online floorplan ray extraction for real-time coarse retrieval.  After obtaining floorplan-side rays, we rank sampled hypotheses by ray consistency:
%\vspace{-0.25em}
%by their consistency with image-side rays:
%$\mathcal{P}_{t,j}=\exp\!\left(-\left\|\bar{\mathbf{r}}_{t,j}^{\mathcal{M}}-\widehat{\bar{\mathbf{r}}}_{t}^{I}\right\|_1/\lambda\right)$,
\begin{equation}
\mathcal{P}_{t,j}
=
\exp \left(
-\frac{
\left\|
\bar{\mathbf{r}}_{t,j}^{\mathcal{M}} - 
\widehat{\bar{\mathbf{r}}}_{t}^{I}
\right\|_1
}{\lambda}
\right),
\label{eq:rank}
\end{equation}
%where $\bar{\mathbf{r}}_{t,j}^{\mathcal{M}}$ and $\widehat{\bar{\mathbf{r}}}_{t}^{I}$ denote the normalized floorplan-side and image-side ray descriptors, respectively, and $\lambda$ controls the score sharpness.
%We retain the top-$K$ hypotheses $\{\mathbf{s}_{t,k}^{c}\}_{k\in\mathcal{K}_t}$ and their queried floorplan-side rays $\{\mathbf{r}_{t,k}^{\mathcal{M}}\}_{k\in\mathcal{K}_t}$ as coarse anchors for subsequent pose refinement.
where $\bar{\mathbf{r}}_{t,j}^{\mathcal{M}}$ denotes the scalar floorplan-side 
rays queried at the $j$-th candidate pose, 
$\widehat{\bar{\mathbf{r}}}_{t}^{I}$ denotes the predicted image-side scalar 
rays, and $\lambda$ controls the score sharpness.
We retain the top-$K$ hypotheses for subsequent pose refinement as 
$\big(\hat{\mathbf{r}}_{t}^{I}, 
\{(\mathbf{s}_{t,k}^{c}, \mathbf{r}_{t,k}^{\mathcal{M}})\}_{k\in\mathcal{K}_t}\big)$,
where $\mathcal{K}_t$ denotes the selected top-$K$ index set and 
$\mathbf{s}_{t,k}^{c}$ is the $k$-th coarse pose anchor.

\textbf{Diffusion-Aided Pose Refinement.} 
%This stage turns the sparse coarse anchors into accurate pose hypotheses. We formulate the refinement of the coarse top-$K$ hypotheses as a residual pose optimization problem. 
After coarse retrieval, the top-$K$ anchors provide plausible but still discretized pose hypotheses.
To move beyond the coarse sampling resolution, we propose a diffusion-aided pose refinement module that formulates refinement as residual pose optimization and produces accurate continuous pose estimates.
For notational clarity, we omit the frame index 
$t$ in this section. 
For the $k$-th coarse pose candidate 
$\mathbf{s}^{c}_{k}\in\mathrm{SE}(2)$, we estimate a relative correction 
$\Delta \mathbf{T}_{k}\in\mathrm{SE}(2)$ conditioned on the image-side ray 
observation $\hat{\mathbf{r}}^{I}$ and the corresponding floorplan-side ray 
observation $\mathbf{r}_{k}^{\mathcal{M}}$. The refined pose is obtained as 
\begin{equation}
\mathbf{s}^{f}_{k}
=
\mathbf{s}^{c}_{k}
\oplus
\Delta \mathbf{T}_{k},
\label{eq:refine_pose_composition}
\end{equation}
where $\oplus$ denotes pose composition in $\mathrm{SE}(2)$.
Inspired by recent optimization methods~\cite{wang2023posediffusion,li2024diffloc,zhang2023genpose}, we cast residual pose estimation as a conditional 
diffusion process rather than one-shot regression, as iterative denoising 
naturally mirrors pose optimization~\cite{wang2023posediffusion}. %Specifically, the residual correction is progressively refined under the guidance of paired image-side and floorplan-side ray observations. Accordingly, 
The residual correction is therefore progressively refined under the guidance of paired image-side and floorplan-side ray observations. We model
the conditional residual distribution as
$p_{\theta}(\mathbf{x}_{0,k}\mid \hat{\mathbf{r}}^{I}, \mathbf{r}_{k}^{\mathcal{M}})$,
where $\mathbf{x}_{0,k}$ parameterizes the residual pose in the local 
coordinate frame of $\mathbf{s}^{c}_{k}$.
%. Specifically, we represent the residual as 
We encode 
the residual as $\mathbf{x}_{0,k}=
[\Delta x_{k},\,\Delta y_{k},\,\cos\Delta\phi_{k},\,\sin\Delta\phi_{k}]^{\top}$, where $(\Delta x_{k},\Delta y_{k})$ denotes the residual translation and 
$\Delta \phi_{k}$ denotes the residual yaw angle. The sine-cosine representation avoids discontinuities caused by angular wrap-around.
During training, the ground-truth residual $\mathbf{x}_{0,k}$ is computed 
from the relative transformation between the coarse pose $\mathbf{s}^{c}_{k}$ 
and the ground-truth pose. We then apply the standard 
Gaussian forward diffusion process~\cite{ho2020denoising}:
\begin{equation}
\mathbf{x}_{\tau,k}
=
\sqrt{\bar{\alpha}_{\tau}}\mathbf{x}_{0,k}
+
\sqrt{1-\bar{\alpha}_{\tau}}\boldsymbol{\epsilon},
\qquad
\boldsymbol{\epsilon}\sim\mathcal{N}(\mathbf{0},\mathbf{I}),
\label{eq:refine_forward_diffusion}
\end{equation}
where $\tau$ denotes the diffusion timestep and $\bar{\alpha}_{\tau}$ is the 
cumulative noise schedule.
To condition the denoising process on the paired ray observations, we first encode the image-side and floorplan-side rays with a shared PointNet-style~\cite{qi2017pointnet} ray encoder $E_{\mathrm{ray}}(\cdot)$. We then introduce 
a Ray Discrepancy Encoder (RDE), denoted as $\Phi(\cdot)$, to compute soft correspondences between the paired ray features and aggregate them into a global conditioning feature: 
$\mathbf{z}_k =
\Phi\!\left(
E_{\mathrm{ray}}(\hat{\mathbf r}^{I}),
E_{\mathrm{ray}}(\mathbf r_k^{M}),
\hat{\mathbf r}^{I},
\mathbf r_k^{M}
\right)$,
where $\mathbf{z}_{k}$ encodes correspondence-aware geometric cues between 
the image-predicted rays and the floorplan rays. Further details of the RDE architecture are provided in Appendix~\ref{supp:rde_details}.
The conditional denoiser 
$D_{\theta}$ takes the noisy residual $\mathbf{x}_{\tau,k}$, the timestep 
$\tau$, and the conditioning feature $\mathbf{z}_{k}$ as input, and is trained 
to predict the injected noise:
\begin{equation}
\mathcal{L}_{\mathrm{diff}}
=
\mathbb{E}_{\tau,\mathbf{x}_{0,k},\boldsymbol{\epsilon}}
\left[
\left\|
D_{\theta}
\left(
\mathbf{x}_{\tau,k},
\tau,
\mathbf{z}_{k}
\right)
-
\boldsymbol{\epsilon}
\right\|_{2}^{2}
\right].
\label{eq:refine_diffusion_loss}
\end{equation}

At inference time, we initialize 
$\mathbf{x}_{T,k}$ from pure noise and perform 
DDIM~\cite{song2020denoising} sampling conditioned on $\mathbf{z}_{k}$. At each reverse 
denoising step, the residual is updated as 
\begin{equation}
\begin{aligned}
\hat{\mathbf{x}}_{\tau-1,k}
=
&\sqrt{\bar{\alpha}_{\tau-1}}
\left(
\frac{
\hat{\mathbf{x}}_{\tau,k}
-
\sqrt{1-\bar{\alpha}_{\tau}}\,
D_{\theta}
}{
\sqrt{\bar{\alpha}_{\tau}}
}
\right)
+
\sqrt{1-\bar{\alpha}_{\tau-1}}\,
D_{\theta}.
\end{aligned}
\label{eq:refine_ddim_sampling}
\end{equation}
After iterating from $\tau=T$ to $\tau=1$, the denoised residual 
$\hat{\mathbf{x}}_{0,k}$ is mapped to an $\mathrm{SE}(2)$ correction 
$\Delta\hat{\mathbf{T}}_{k}$ and composed with the coarse hypothesis to obtain 
the refined pose in Eq.~\ref{eq:refine_pose_composition}, yielding a locally 
consistent pose estimate guided by paired ray geometry.

\textbf{Training and Inference.}
During training, we additionally sample a co-visible \textit{gold anchor} near 
the ground truth to provide meaningful residual corrections and stabilize 
geometry-guided refinement. At inference, no gold anchor is used; all top-$K$ 
anchors from coarse retrieval are refined in batch, re-queried online, and 
re-scored by Eq.~\ref{eq:rank} to select the final pose. In our experiments, we use a fixed budget of $2000$ coarse sampled locations based on the average floorplan scale, with a rotation interval of $15^\circ$, 
and retain the top $K=10$ anchors for refinement. All experiments are conducted on the NVIDIA RTX 4090 GPU.
% For training the diffusion-aided pose refinement module, we additionally sample
% an auxiliary \textit{gold anchor} around the ground-truth pose. This anchor is
% selected to maintain sufficient co-visibility with the target pose, thereby providing reliable ray-level correspondences between floorplan-side and image-side observations. Such training anchors provide meaningful residual corrections and stabilize 
% geometry-guided pose refinement.
% At inference time, no \textit{gold anchor} is used. Given the top-$K$ anchors 
% retrieved by the coarse stage, we refine all anchors in batch using the current 
% %image 
% image-side
% rays. For each refined pose, we re-query its 
% %floorplan 
% floorplan-side
% rays online and score the resulting candidate using 
% Eq.~\ref{eq:rank}. The candidate with the highest score is selected as the final pose estimate. In our experiments, we use a fixed budget of $1000$ coarse sampled locations based on the average floorplan scale, with a rotation interval of $15^\circ$, 
% and retain the top $K=10$ anchors for refinement. All experiments are conducted on an NVIDIA RTX 4090 GPU.

\subsection{Sequential Localization}
\label{sec:seq_loc}
%For sequential localization, our histogram filter~\cite{thrun2002probabilistic} follows the transition model of \fthreeloc~\cite{chen2024f3loc}, but differs in the observation likelihood construction. Instead of densely evaluating likelihoods from an offline ray database, 
For robots with sequential RGB observations, sequential localization aims to maintain temporally consistent pose estimates.
To bridge single-frame localization and temporal fusion, we construct observation likelihoods online from the sparse outputs of our coarse-to-fine estimator: coarse candidates provide broad pose-space support, while refined hypotheses sharpen high-confidence regions.
The resulting likelihood is fused over time with a histogram filter~\cite{thrun2002probabilistic}.

%Specifically, we first initialize a sparse likelihood from online coarse pose samples and update its top-$K$ peak regions using the corresponding refined poses and confidence scores. 
Specifically, we first scatter the online coarse pose samples into the
discretized $\mathrm{SE}(2)$ pose volume according to their matching scores,
forming an initial sparse likelihood. The top-$K$ peak regions are then refined
by our diffusion-aided pose refinement module, and their likelihood values are
updated using the refined poses and confidence scores. The remaining sampled
locations keep their initial coarse likelihoods. 
Finally, the updated sparse pose-score set $\mathcal{A}_t$ is interpolated by 
$\mathcal{I}(\cdot)$ and spatially smoothed by $\mathcal{G}_{\sigma}$ to obtain 
$p(o_t \mid s_t)=\left[\mathcal{G}_{\sigma} *
\mathcal{I}(\mathcal{A}_t)\right](s_t)$.
% Finally,
% the updated sparse 
% likelihood is then interpolated and spatially smoothed to obtain the dense 
% single-frame observation likelihood $p(o_t \mid s_t)=\left[\mathcal{G}_{\sigma} * \mathcal{I}(\mathcal{A}_t)\right](s_t)$, 
% where $\mathcal{A}_t$ denotes the updated sparse pose-score set, 
% $\mathcal{I}(\cdot)$ is the interpolation operator, and 
% $\mathcal{G}_{\sigma}$ is the spatial smoothing kernel.
This construction preserves 
the broad support of coarse retrieval while injecting more reliable likelihoods 
around refined high-confidence regions. 
Given the observation likelihood $p(o_t \mid s_t)$, we follow the histogram 
filtering formulation to perform sequential localization over the discretized 
pose space. The posterior distribution from the previous frame is first 
propagated by the inter-frame ego motion $\mathbf{u}_t$, yielding the prior 
distribution at the current frame:
\vspace{-0.3em}
\begin{equation}
p(s_t \mid o_{1:t-1}, \mathbf{u}_{1:t})
=
\sum_{s_{t-1}\in\mathcal{S}}
p(s_t \mid s_{t-1}, \mathbf{u}_t)\,
p(s_{t-1} \mid o_{1:t-1}, \mathbf{u}_{1:t-1}).
\label{eq:seq_prediction}
\end{equation}
The propagated distribution is then updated with the current observation 
likelihood to produce the posterior distribution at the current frame:
\vspace{-0.3em}
\begin{equation}
p(s_t \mid o_{1:t}, \mathbf{u}_{1:t})
=
\eta\,
p(o_t \mid s_t)\,
p(s_t \mid o_{1:t-1}, \mathbf{u}_{1:t}),
\label{eq:seq_update}
\end{equation}
where $\eta$ is a normalization factor. 
%Following \cite{chen2024f3loc}, \zp{Do not always metion f3loc}
We implement the transition model 
$p(s_t \mid s_{t-1}, \mathbf{u}_t)$ with decoupled translation and rotation 
filters over the pose volume to accelerate the filtering update. At each time 
step, the final pose is obtained as the maximum a posteriori estimate.

\section{Experiments}
\label{sec:exp}

\subsection{Experimental Setup}

\textbf{Dataset.} We consider two widely used datasets with different %data 
distributions: Gibson~\cite{shen2021igibson} and Structured3D~\cite{zheng2020structured3d}. 
Gibson contains 118 indoor scenes with floorplans and monocular images %RGB observations 
captured with a $108^\circ$ FoV, %field of view, 
collected by \fthreeloc~\cite{chen2024f3loc}. It includes Gibson(f), %consisting of 
with 24,779 four-view sequences for single-frame localization, and Gibson(t), with %consisting of 
118 trajectories with 280 to 5,152 steps for sequential localization. Structured3D 
contains 3,296 %fully 
furnished indoor environments and 78,453 perspective 
images captured with an $80^\circ$ horizontal field of view. All evaluations follow official train/test splits.

% \subsection{Baseline and Metrics}

\textbf{Baseline and Metrics.} We compare our method with representative floorplan localization approaches 
from different categories. \fthreeloc~\cite{chen2024f3loc} is used as the primary baseline, as it is a 
strong state-of-the-art depth-based method, and we adopt its monocular version 
for comparison. 
%\hc{We additionally consider UnLoc~\cite{wuest2025unloc}, which
% builds upon F$^3$Loc with stronger foundation-model-based depth estimation. A
% controlled comparison with UnLoc, including evaluation under matched depth
% predictions, is provided in the supplementary material.}
%We additionally include UnLoc~\cite{wuest2025unloc}, which builds upon F$^3$Loc with stronger foundation-model-based depth estimation. To disentangle the effect of depth quality from the localization pipeline, we replace our default F$^3$Loc depth predictions with UnLoc depth predictions at test time, without retraining the diffusion refinement model. This provides a depth-matched comparison with UnLoc and evaluates whether FreeLoc can consistently benefit from stronger depth estimates in a plug-and-play manner.
We also evaluate against PF-Net~\cite{pf-net} and LASER~\cite{semantic-based:min2022laser}, two representative non-depth-based methods for differentiable particle-filter-based and feature-based floorplan 
localization, respectively.
Following prior work~\cite{chen2024f3loc, semantic-based:min2022laser}, for single-frame localization, we 
report recall at translation thresholds of $0.1$m, $0.5$m, and $1.0$m, 
together with recall under the joint pose threshold of $1.0$m and 
$30^\circ$. For sequential localization, we report success rate and tracking accuracy. 
A trajectory is considered successful at threshold $X$ m if the predictions 
stay within $X$ m of the ground truth over the last 10 frames. We also 
report RMSE over the last 10 frames for both successful and all trajectories. Following F3Loc, our default sequential evaluation uses GT-derived inter-frame ego-motion for controlled comparison; robustness to noisy and temporarily unavailable ego-motion is evaluated in
Sec.~4.5.

\begin{table*}[t]
\centering
\caption{
Single-frame localization recall (\%) on Gibson(f) and Structured3D.
Bold denotes the best result.
}
\label{tab:single_frame_results}

\scalebox{0.78}{%
\begin{minipage}{1.0\textwidth}
\centering

\small
\setlength{\tabcolsep}{4.2pt}
\renewcommand{\arraystretch}{1.08}

\begin{tabular}{lcccccccc}
\toprule
\multirow{2}{*}{\textbf{Method}}
& \multicolumn{4}{c}{\textbf{Gibson(f)}}
& \multicolumn{4}{c}{\textbf{Structured3D}} \\
\cmidrule(lr){2-5}
\cmidrule(lr){6-9}

& $@0.1\mathrm{m}\uparrow$
& $@0.5\mathrm{m}\uparrow$
& $@1\mathrm{m}\uparrow$
& $@1\mathrm{m}30^\circ\uparrow$
& $@0.1\mathrm{m}\uparrow$
& $@0.5\mathrm{m}\uparrow$
& $@1\mathrm{m}\uparrow$
& $@1\mathrm{m}30^\circ\uparrow$ \\
\midrule

PF-Net
& 0
& 1.0
& 4.1
& 1.1
& 0.1
& 1.2
& 4.4
& 1.4 \\

LASER
& 0.2
& 3.9
& 9.8
& 6.2
& 0.5
& 5.2
& 10.0
& 7.8 \\

F$^3$Loc
& 5.3
& 30.7
& 37.9
& 36.3
& 1.7
& 15.0
& 23.0
& 21.9 \\

\textbf{Ours}
& \textbf{10.0}
& \textbf{39.4}
& \textbf{44.6}
& \textbf{42.9}
& \textbf{3.1}
& \textbf{23.9}
& \textbf{30.5}
& \textbf{29.1} \\

\bottomrule
\end{tabular}

\end{minipage}%
}

\vspace{-1.8em}
\end{table*}
\subsection{Single Frame Localization Results}

\textbf{Results on Gibson(f) and Structured3D.}
As shown in Table~\ref{tab:single_frame_results}, our FreeLoc consistently
outperforms all baselines on both Gibson(f) and Structured3D. Compared with the
strongest baseline F$^3$Loc, ours improves all recall metrics by
$4.7$--$8.7$ percentage points on Gibson(f) and $1.4$--$8.9$ percentage points
on Structured3D, demonstrating consistent gains across different data
distributions. We attribute these improvements mainly to the proposed pose
refinement stage. Unlike offline database matching, whose accuracy is
constrained by the discretization of pre-sampled floorplan poses, our refinement
module optimizes coarse anchors in a continuous pose space, allowing predictions
to move beyond sampled locations and achieve finer pose accuracy. 
Moreover, while both F$^3$Loc and our method benefit from floorplan-ray
representations and substantially outperform PF-Net and LASER, FreeLoc further
improves upon F$^3$Loc by replacing dense offline ray matching with online
coarse sampling and continuous pose refinement.

\subsection{Sequential Localization Results}
% \begin{table}[t]
% \centering
% \caption{Sequential localization results on Gibson(t).}
% \label{tab:sequential_results}
% \begin{tabular}{lccc}
% \toprule
% & $\mathrm{SR}@1\mathrm{m}$ (\%) $\uparrow$
% & $\mathrm{RMSE}_{\mathrm{succ.}}$ $\downarrow$
% & $\mathrm{RMSE}_{\mathrm{all}}$ $\downarrow$ \\

% \midrule
% PF-net
% & 8
% & 0.43
% & 4.27
% \\

% LASER
% & 35
% & 0.31
% & 2.69\\

% $\mathrm{F}^{3}\mathrm{Loc}$ 
% & 89.2
% & 0.18
% & 0.88 \\

% Ours
% & \textbf{97.3}
% & \textbf{0.17}
% & \textbf{0.24}\\
% \bottomrule
% \end{tabular}
% \end{table}

% \begin{wraptable}{r}{0.50\textwidth}
% \vspace{-1.5em}
% \centering
% \caption{Sequential %localization 
% results on Gibson(t).}
% \label{tab:sequential_results}
% \scriptsize
% \setlength{\tabcolsep}{2.5pt}
% \resizebox{\linewidth}{!}{
% \begin{tabular}{lccc}
% \toprule
% & $\mathrm{SR}@1\mathrm{m}$ (\%) $\uparrow$
% & $\mathrm{RMSE}_{\mathrm{succ}}$ $\downarrow$
% & $\mathrm{RMSE}_{\mathrm{all}}$ $\downarrow$ \\
% \midrule
% PF-Net
% & 8.0
% & 0.43
% & 4.27 \\

% LASER
% & 35.0
% & 0.31
% & 2.69 \\

% F$^3$Loc
% & 86.5
% & \textbf{0.14}
% & 0.87 \\

% Ours
% & \textbf{100}
% & 0.15
% & \textbf{0.15} \\
% \bottomrule
% \end{tabular}
% }
% \vspace{-1.5em}
% \end{wraptable}

\begin{wraptable}{r}{0.48\textwidth}
\vspace{-1.5em}
\centering
\caption{
Sequential localization results on Gibson(t).
}
\label{tab:sequential_results}

\scriptsize
\setlength{\tabcolsep}{2.5pt}
\renewcommand{\arraystretch}{1.03}

\resizebox{\linewidth}{!}{
\begin{tabular}{lccc}
\toprule
\textbf{Method}
& $\mathrm{SR}@1\mathrm{m}$ (\%) $\uparrow$
& $\mathrm{RMSE}_{\mathrm{succ}}$ (m) $\downarrow$
& $\mathrm{RMSE}_{\mathrm{all}}$ (m) $\downarrow$ \\
\midrule

PF-Net
& 8.0
& 0.43
& 4.27 \\

LASER
& 35.0
& 0.31
& 2.69 \\

F$^3$Loc
& 86.5
& \textbf{0.14}
& 0.87 \\

\textbf{Ours}
& \textbf{100.0}
& 0.15
& \textbf{0.15} \\

\bottomrule
\end{tabular}
}

\vspace{-1.8em}
\end{wraptable}

\textbf{Results on Gibson(t).}
Table~\ref{tab:sequential_results} reports the sequential localization results
on Gibson(t), where the sequence length is set to 100 frames. Our method
achieves the best success rate and overall tracking accuracy. Compared with
F$^3$Loc, FreeLoc improves the success rate by $13.5$ percentage points and
reduces RMSE$_{\rm all}$ by $82.8\%$, demonstrating substantially more robust
tracking over long trajectories. For RMSE$_{\rm succ}$, FreeLoc remains
comparable to F$^3$Loc, with only a marginal difference ($0.15\,\mathrm{m}$ vs.\
$0.14\,\mathrm{m}$), indicating that both methods achieve similarly low errors
once successfully localized. This slight gap mainly stems from reconstructing
dense observation likelihoods from sparse online candidates, which may slightly
reduce local precision compared with dense likelihood evaluation over a
precomputed ray database. Nevertheless, by initializing observation likelihoods
from online sampled candidates and updating them with refined pose hypotheses,
FreeLoc provides reliable frame-level evidence for histogram filtering, enabling
stable sequential localization without an offline ray database. Detailed
qualitative comparisons are provided in Appendix~\ref{supp:sequential_qualitative}.

\textbf{Timing.} We report the average per-frame inference time under the Gibson(t) sequential localization setting, together with the scene-specific offline sampling cost required by database-based methods. The offline sampling time is measured on the first Gibson(t) test scene, which covers approximately $10\,\mathrm{m}\times6\,\mathrm{m}$. Following the official \fthreeloc~implementation, we use its released sampling code and testing-time sampling resolution to construct the offline ray database. \fthreeloc~achieves fast inference~($0.014\,\mathrm{s}$ per frame) once the scene-specific ray database has been precomputed, but constructing this database takes $1.48$ hours for a single scene. In contrast, our method localizes directly from the provided floorplan without offline ray-database construction, while still running about $15$ Hz ($0.069\,\mathrm{s}$ per frame) including online sampling. Therefore, when deployed in a new scene from 
scratch, our unified coarse-to-fine framework substantially reduces the 
overall localization cost by eliminating the time-consuming scene-specific preprocessing stage.

\subsection{Real-World Experiment}
\begin{figure}[t]
    \vspace{-0.5em}
    \centering
    \includegraphics[width=0.8\linewidth]{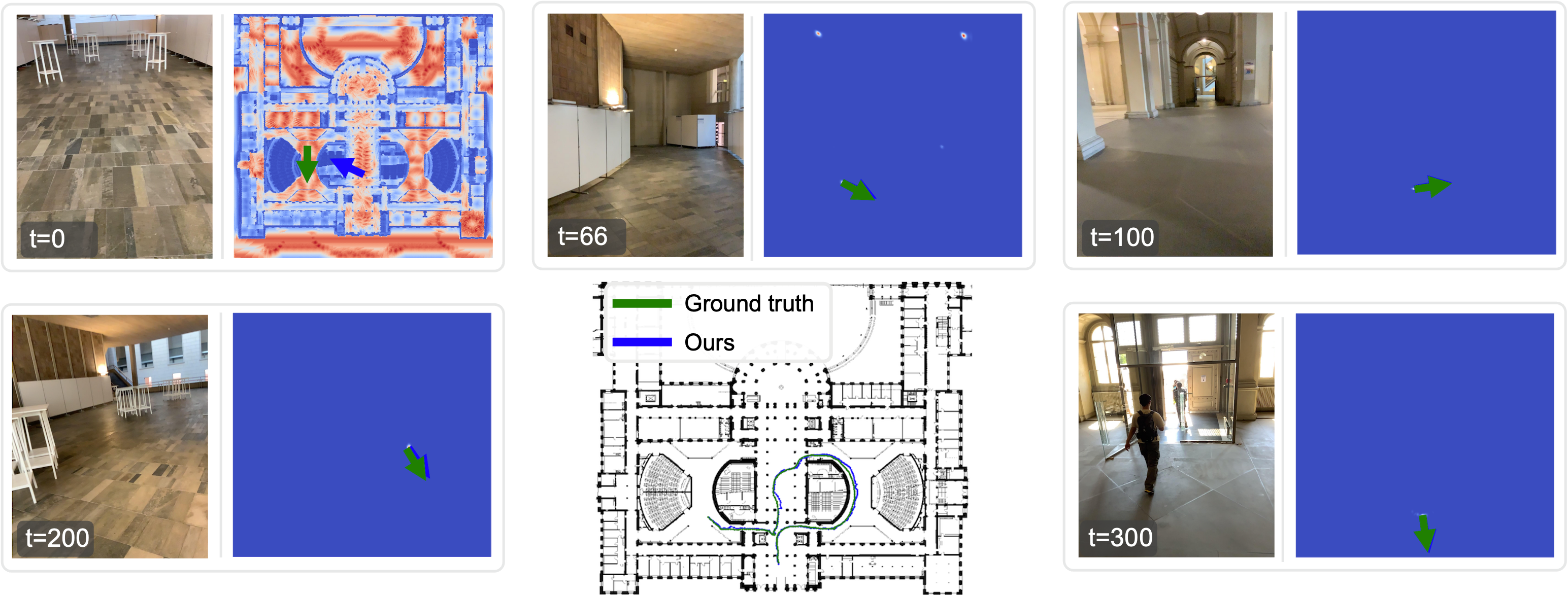}
    \vspace{-0.8em}
    \caption{\textbf{Real-world performance of \ours.} The center panel shows the floorplan with converged predicted and ground-truth 
trajectories; surrounding panels show time-ordered RGB observations and 
posterior probabilities. Blue indicates predictions and green indicates ground 
truth.}
    \label{fig:real_world}
    \vspace{-1.5em}
\end{figure}

To further validate real-world deployability, we build a real-world setting from the HGE building 
in the LaMAR dataset~\cite{sarlin2022lamar}, since no public 
real-world indoor dataset provides both continuous RGB sequences and 
corresponding floorplan maps under our target setting.
HGE is a 
large-scale indoor scene with rooms, corridors, and open halls, covering about 
$130\,\mathrm{m}\times130\,\mathrm{m}$. The sequences are captured by a 
handheld iPhone with a narrow $48^\circ$ FoV. As shown in Fig.~\ref{fig:real_world}, the posterior distribution is initially 
multi-modal due to the large search space and the ambiguity of narrow-FoV 
observations. \ours gradually concentrates the posterior around the 
correct region and forms a stable, sharp peak within about $66$ frames. After 
convergence, the estimated trajectory remains closely aligned with the ground 
truth throughout the sequence, demonstrating the stable long-term localization 
capability of \ours in a large and structurally complex real-world 
environment. In terms of efficiency, \ours enables faster scene preparation and deployment
than \fthreeloc, while still maintaining an inference speed of approximately
3 Hz on this large-scale indoor floorplan. 
Additional qualitative results are available on our project page:~\url{https://zju3dv.github.io/freeloc/}.

% \begin{figure}[t]
%     \centering
%     \includegraphics[width=\linewidth]{figure/realworld_exp.png}
%     \caption{\textbf{Real-world performance of \ours on the HGE building.} The center panel shows 
% the official HGE floorplan and the trajectory tracked by our method after 
% successful convergence. The two side columns present the corresponding RGB 
% observations and predicted posterior probability maps at selected frames. Blue 
% denotes the predicted poses/trajectory, and green denotes the ground truth.}
%     \label{fig:real_world}
% \end{figure}

%Consider moving this to the appendix
%
\subsection{Ablation Studies}
% We conduct an ablation study to evaluate the contribution of each stage in our 
% coarse-to-fine localization framework. \textit{Stage-I} denotes online coarse pose retrieval, 
% \textit{Stage-I+ICP} refines the coarse anchors with a classical ICP-style 
% alignment, and \textit{Stage-II} applies our diffusion-aided pose refinement.
% As shown in Table~\ref{tab:ablation_refine}, Stage-I already provides 
% reasonable localization performance, indicating that online floorplan querying 
% can retrieve plausible coarse pose anchors. Adding ICP improves recall 
% across all thresholds, showing that local geometric pose optimization is 
% beneficial. However, our Stage-II further outperforms the ICP variant, 
% especially under the strict $0.1\,\mathrm{m}$ threshold. This demonstrates that 
% the proposed diffusion-aided refinement is more effective than deterministic 
% local alignment for optimizing coarse anchors under noisy and partial 
% ray-level observations, leading to more accurate fine-grained localization.

% \begingroup
% \setlength{\intextsep}{0pt}
% \setlength{\columnsep}{1em}

\begin{wraptable}{r}{0.48\textwidth}
\vspace{-1.5em}
\centering
\caption{Pose refinement under different coarse-initialization quality.}
\label{tab:refinement_error}

\scriptsize
\setlength{\tabcolsep}{5.0pt}
\renewcommand{\arraystretch}{1.05}

\begin{tabular}{lcc}
\toprule
\textbf{Initial error}
& \textbf{Before (m)}
& \textbf{After (m)} \\
\midrule

$\leq 0.5\,\mathrm{m}$
& 0.25
& 0.18 \\

$\leq 1.0\,\mathrm{m}$
& 0.35
& 0.25 \\

$\leq 1.5\,\mathrm{m}$
& 0.42
& 0.32 \\

All
& 3.14
& 3.09 \\

\bottomrule
\end{tabular}

\vspace{-1.5em}
\end{wraptable}
\textbf{Coarse-to-Fine Localization Analysis.}
We analyze the behavior of \emph{Diffusion-Aided Pose Refinement} under
different qualities of coarse initialization. As shown in
Table~\ref{tab:refinement_error}, when coarse hypotheses lie within
$0.5\,\mathrm{m}$, $1.0\,\mathrm{m}$, and $1.5\,\mathrm{m}$ of the ground
truth, diffusion refinement consistently reduces the average translation
error, demonstrating effective residual pose correction from plausible
initializations. In contrast, when all coarse hypotheses are considered
regardless of their initialization quality, the improvement is marginal
($3.14\,\mathrm{m}$ to $3.09\,\mathrm{m}$). This limited gain arises mainly
because the ``All'' set contains many poorly initialized hypotheses, including
cross-room cases that are geometrically far from the ground-truth pose. Such
hypotheses fall outside the effective range of local residual correction and
are therefore difficult to recover through refinement alone. These results
clarify the intended division of labor in our coarse-to-fine framework:
\emph{Online Coarse Pose Retrieval} narrows the global search to geometrically
plausible pose hypotheses, while \emph{Diffusion-Aided Pose Refinement}
performs precise local residual correction to further improve localization
accuracy.

\textbf{Diffusion-Aided Pose Refinement  vs. ICP Optimization.}
We further compare our diffusion-aided pose refinement with ICP optimization,
a representative deterministic alternative for residual geometric alignment,
to isolate the benefit of learned diffusion-based optimization. As shown in
Table~\ref{tab:refine_icp_seq}, diffusion refinement improves sequential
SR@1m and SR@0.5m by $2.7$ percentage points, with a larger gain of 8.2 percentage points at the stricter
$0.2\,\mathrm{m}$ threshold, while reducing RMSE$_{\rm all}$ by $28.6\%$.
These gains are achieved with only $6.8\,\mathrm{ms}$ additional latency
($35.2\,\mathrm{ms}$ vs.\ $28.4\,\mathrm{ms}$ for ICP optimization),
demonstrating that our learned diffusion-based optimizer provides more accurate pose
refinement than deterministic ICP with only modest computational overhead.

\begin{table}[t]
\centering
\caption{Sequential localization comparison between ICP optimization and diffusion-aided pose refinement on Gibson(t).}
\label{tab:refine_icp_seq}

\vspace{-0.5em}

\scriptsize
\setlength{\tabcolsep}{2.8pt}
\renewcommand{\arraystretch}{1.05}

\resizebox{0.8\linewidth}{!}{%
\begin{tabular}{lcccc}
\toprule
\textbf{Refinement}
& \textbf{SR@1m (\%)} $\uparrow$
& \textbf{SR@0.5m (\%)} $\uparrow$
& \textbf{SR@0.2m (\%)} $\uparrow$
& \textbf{RMSE$_{\rm all}$ (m)} $\downarrow$ \\
\midrule

ICP Optimization
& 97.3
& 97.3
& 43.2
& 0.21 \\

Diffusion-Aided Pose Refinement
& \textbf{100.0}
& \textbf{100.0}
& \textbf{51.4}
& \textbf{0.15} \\

\bottomrule
\end{tabular}%
}

\vspace{-2.0em}
\end{table}
\noindent\textbf{Robustness to Ego-Motion Uncertainty.}
Following \fthreeloc, our default sequential evaluation uses GT-derived
inter-frame ego-motion for controlled comparison. We further evaluate sequential
localization robustness under three settings. The first two emulate common
ego-motion errors: (i) frame-wise Gaussian noise models random motion-estimation
errors, while (ii) motion-dependent noise increases with motion magnitude, with
$\sigma_t=b_t+a_td_t$ and
$\sigma_r=b_r+a_rd_t+c_rd_r$. The third setting models temporary ego-motion
failure: (iii) ego-motion is unavailable for five consecutive frames, during
which the last valid relative motion is reused under a constant-motion
assumption. As shown in Table~5, \ours maintains strong sequential performance under both
noisy and temporarily unavailable ego-motion, demonstrating robust sequential
localization under representative ego-motion disturbances.
\begin{table}[!htbp]
\centering
\caption{Robustness to inter-frame ego-motion uncertainty on Gibson(t).}
\label{tab:ego_motion_robustness}
\vspace{-0.5em}

\resizebox{0.82\linewidth}{!}{%
\begin{minipage}{\linewidth}

\footnotesize
\renewcommand{\arraystretch}{0.98}

% ============================================================
% (a) Ego-motion robustness
% ============================================================
\begin{minipage}[t]{0.59\linewidth}
\vspace{0pt}
\centering
\textbf{(a) Ego-motion robustness}\strut\par
\vspace{0.20em}

\setlength{\tabcolsep}{2.8pt}
\begin{tabular}{@{}llccc@{}}
\toprule
\textbf{Setting}
& \textbf{Level}
& \textbf{SR@1m}
& \textbf{RMSE$_{\rm succ}$}
& \textbf{RMSE$_{\rm all}$} \\
\midrule

Clean
& -- & 100.0 & 0.145 & 0.145 \\

\multirow{3}{*}{Gaussian}
& Mild     & 100.0 & 0.147 & 0.147 \\
& Moderate & 100.0 & 0.171 & 0.171 \\
& Severe   & 97.3  & 0.235 & 0.307 \\

\multirow{3}{*}{Motion-dep.}
& Mild     & 100.0 & 0.145 & 0.145 \\
& Moderate & 100.0 & 0.150 & 0.150 \\
& Severe   & 100.0 & 0.163 & 0.163 \\

Ego-motion unavail.
& 5 frames & 100.0 & 0.158 & 0.158 \\

\bottomrule
\end{tabular}
\end{minipage}%
\hfill
% ============================================================
% (b) Noise parameters
% ============================================================
\begin{minipage}[t]{0.39\linewidth}
\vspace{0pt}
\centering
\textbf{(b) Noise parameters}\strut\par
\vspace{0.20em}

\setlength{\tabcolsep}{1.6pt}
\begin{tabular}{@{}llccc@{}}
\toprule
\textbf{Noise}
& \textbf{Param.}
& \textbf{Mild}
& \textbf{Mod.}
& \textbf{Sev.} \\
\midrule

\multirow{2}{*}{Gaussian}
& $\sigma_t$ (m)
& 0.01 & 0.03 & 0.05 \\
& $\sigma_r$ ($^\circ$)
& 0.5 & 1.0 & 2.0 \\

\cmidrule(lr){1-5}

\multirow{5}{*}{Motion-dep.}
& $b_t$ (m)
& 0.005 & 0.01 & 0.02 \\
& $a_t$
& 0.01 & 0.03 & 0.05 \\
& $b_r$ ($^\circ$)
& 0.2 & 0.5 & 1.0 \\
& $a_r$ ($^\circ$/m)
& 0.2 & 0.5 & 1.0 \\
& $c_r$
& 0.01 & 0.03 & 0.05 \\

\noalign{\vskip 5.0pt}

\bottomrule
\end{tabular}
\end{minipage}

\end{minipage}%
}

\vspace{-2.0em}
\end{table}

%===============================================================================

\section{Conclusion}
\label{sec:conclusion}

We presented \ours, an online RGB-based floorplan localization framework that treats the floorplan as a directly queryable geometric map rather than a dense scene-specific offline database. \ours builds a unified online coarse-to-fine pipeline that combines on-the-fly floorplan ray querying with diffusion-aided pose refinement to retrieve plausible anchors and refine them into accurate continuous pose estimates. The same coarse anchors and refined hypotheses are further used to construct online observation likelihoods for sequential localization, bridging single-frame localization and probabilistic temporal fusion without offline databases. Experiments on various datasets demonstrate accurate and efficient localization, as well as immediate deployability in new %indoor 
environments.

\section{Limitations}
\label{sec:limitation}

    Although our method demonstrates strong localization performance and favorable
deployability, its generalization remains limited under substantially different
camera settings or scene distributions. This limitation mainly stems from the
monocular floorplan-depth estimator, which is sensitive to camera intrinsics,
field of view, and scene scale. Moreover, while FreeLoc supports efficient inference in typical indoor
environments, the enlarged search space of very large floorplans can still
affect real-time performance, although FreeLoc retains faster and more
convenient deployment than existing database-based approaches. Future work will explore
more generalizable depth estimation and hierarchical floorplan
querying to improve both cross-domain robustness and runtime scalability.
%===============================================================================

\clearpage
% The acknowledgments are automatically included only in the final and preprint versions of the paper.
\acknowledgments{We thank Yanqi Zhu for help with processing the Structured3D dataset. This work was partially supported by the NSFC (No.~62572425 and No.~624B2132).}

%===============================================================================

\bibliography{example}  % .bib

\clearpage
\appendix
\section{Additional Implementation Details}
\label{supp:implementation_details}  

\subsection{Training Details}
\label{supp:training_details}

For the floorplan-depth estimator, we follow \fthreeloc~\cite{chen2024f3loc}~and train the model for 
$100$ epochs on both Gibson~\cite{shen2021igibson} and Structured3D~\cite{zheng2020structured3d} with a batch size of $32$. For the 
diffusion-aided pose refinement network, we train the model for $200$ epochs on 
each dataset with a batch size of $128$. We use the AdamW optimizer~\cite{adamw} with an 
initial learning rate of $1\times10^{-3}$, together with a cosine annealing 
learning-rate scheduler. We adopt a cosine noise schedule during training and use $10$-step DDIM
sampling~\cite{song2020denoising} at inference. The refinement network contains
approximately $3.0$M parameters.

\subsection{Ray Discrepancy Encoder Details}
\label{supp:rde_details}

Given the image-side ray observation $\hat{\mathbf{r}}^{I}$ and the
floorplan-side ray observation $\mathbf{r}_{k}^{\mathcal{M}}$ associated with
the $k$-th coarse anchor, both represented as ray-indexed 2D point sets in
$\mathbb{R}^{V_r\times 2}$, we first extract point-wise ray features using a
shared PointNet-style~\cite{qi2017pointnet} ray encoder:
\begin{equation}
\mathbf{F}^{I}
=
E_{\mathrm{ray}}(\hat{\mathbf{r}}^{I}),
\qquad
\mathbf{F}_{k}^{\mathcal{M}}
=
E_{\mathrm{ray}}(\mathbf{r}_{k}^{\mathcal{M}}),
\label{eq:app_rde_ray_encoder}
\end{equation}
where $\mathbf{F}^{I},\mathbf{F}_{k}^{\mathcal{M}}\in\mathbb{R}^{V_r\times D}$.
The shared encoder maps the image-side and floorplan-side ray observations into
a common feature space, enabling their local geometric patterns to be compared
consistently. The Ray Discrepancy Encoder (RDE) is then applied on top of these
point-wise features, together with the original ray coordinates, to construct a
compact geometric condition for diffusion-aided pose refinement.

The goal of RDE is to capture how the image-side ray observation differs from
the floorplan-side ray observation under the current coarse anchor. To establish
flexible associations between the two ray sets, we first project their
point-wise features into a normalized matching space:
\begin{equation}
\bar{\mathbf{F}}^{I}
=
\operatorname{Normalize}\left(\Pi_I(\mathbf{F}^{I})\right),
\qquad
\bar{\mathbf{F}}_{k}^{\mathcal{M}}
=
\operatorname{Normalize}\left(\Pi_{\mathcal{M}}(\mathbf{F}_{k}^{\mathcal{M}})\right),
\label{eq:app_rde_projection}
\end{equation}
where $\Pi_I(\cdot)$ and $\Pi_{\mathcal{M}}(\cdot)$ are learnable projection
heads for the two ray features, and $\operatorname{Normalize}(\cdot)$ denotes
$\ell_2$ normalization along the feature dimension. We then compute the
pairwise similarity matrix as
\begin{equation}
\mathbf{S}_{k}
=
\frac{
\bar{\mathbf{F}}^{I}
\left(\bar{\mathbf{F}}_{k}^{\mathcal{M}}\right)^{\top}
}{\gamma},
\label{eq:app_rde_similarity}
\end{equation}
where $\gamma$ is a temperature parameter. Bidirectional soft correspondence
matrices are obtained by applying row-wise softmax to $\mathbf{S}_{k}$ and its
transpose:
\begin{equation}
\mathbf{A}_{k}^{I\rightarrow\mathcal{M}}
=
\operatorname{softmax}_{\mathrm{row}}(\mathbf{S}_{k}),
\qquad
\mathbf{A}_{k}^{\mathcal{M}\rightarrow I}
=
\operatorname{softmax}_{\mathrm{row}}(\mathbf{S}_{k}^{\top}).
\label{eq:app_rde_soft_correspondence}
\end{equation}
This soft correspondence preserves the ray-wise structure while relaxing strict index-wise matching. Strict index-wise matching can be unreliable under pose misalignment or image-side ray prediction errors, since the ray indices between the two observations may become shifted and no longer form exact correspondences. By allowing each ray to attend to all candidate rays in the other observation, the proposed soft correspondence enables more flexible geometric matching and improves robustness to imperfect coarse anchors and noisy ray observations.

For the image-to-floorplan direction, we use
$\mathbf{A}_{k}^{I\rightarrow\mathcal{M}}$ to compute the softly matched
floorplan-side feature and coordinate for each image-side ray:
\begin{equation}
\tilde{\mathbf{F}}_{k}^{\mathcal{M}}
=
\mathbf{A}_{k}^{I\rightarrow\mathcal{M}}
\mathbf{F}_{k}^{\mathcal{M}},
\qquad
\tilde{\mathbf{r}}_{k}^{\mathcal{M}}
=
\mathbf{A}_{k}^{I\rightarrow\mathcal{M}}
\mathbf{r}_{k}^{\mathcal{M}}.
\label{eq:app_rde_soft_matched_map}
\end{equation}
We then construct a point-wise discrepancy descriptor by combining
feature-level compatibility and ray-level geometric discrepancy:
\begin{equation}
\begin{aligned}
\mathbf{Q}_{k}^{I\rightarrow\mathcal{M}}
=
\psi\Bigl(
\bigl[
&\mathbf{F}^{I},
\tilde{\mathbf{F}}_{k}^{\mathcal{M}},
\mathbf{F}^{I}-\tilde{\mathbf{F}}_{k}^{\mathcal{M}},
\mathbf{F}^{I}\odot\tilde{\mathbf{F}}_{k}^{\mathcal{M}},
\\
&\hat{\mathbf{r}}^{I},
\tilde{\mathbf{r}}_{k}^{\mathcal{M}},
\hat{\mathbf{r}}^{I}-\tilde{\mathbf{r}}_{k}^{\mathcal{M}},
\hat{\mathbf{r}}^{I}\odot\tilde{\mathbf{r}}_{k}^{\mathcal{M}}
\bigr]
\Bigr),
\end{aligned}
\label{eq:app_rde_pair_descriptor}
\end{equation}
where $\psi(\cdot)$ is a lightweight pair encoder, $\odot$ denotes element-wise
multiplication, and $[\cdot]$ denotes concatenation along the feature dimension.
The feature terms compare each source ray feature with its softly matched
target feature, where the difference term highlights feature mismatch and the
product term provides a multiplicative compatibility cue. The geometric terms
perform the same comparison in the ray coordinate space, where the coordinate
difference directly encodes the residual displacement between the paired ray
observations, while the coordinate product provides an additional interaction
cue. Together, these terms allow the descriptor to encode both compatibility in
the learned geometric feature space and explicit discrepancy in the ray
coordinate space. The floorplan-to-image descriptor
$\mathbf{Q}_{k}^{\mathcal{M}\rightarrow I}$ is constructed symmetrically by
swapping the source and target ray observations.

Finally, RDE aggregates the point-wise discrepancy descriptors into a global
ray discrepancy condition. Since the residual pose is determined by the overall
discrepancy pattern between the two ray observations rather than by an isolated
ray mismatch, we summarize the point-wise descriptors through complementary
global pooling operations. A confidence head predicts a scalar importance score
for each descriptor, which is normalized along the ray dimension:
\begin{equation}
\boldsymbol{\alpha}_{k}^{I\rightarrow\mathcal{M}}
=
\operatorname{softmax}
\left(
h(\mathbf{Q}_{k}^{I\rightarrow\mathcal{M}})
\right).
\label{eq:app_rde_confidence}
\end{equation}
We combine confidence-weighted mean pooling with max pooling:
\begin{equation}
\mathbf{g}_{k,\mathrm{mean}}^{I\rightarrow\mathcal{M}}
=
\sum_{i=1}^{V_r}
\alpha_{k,i}^{I\rightarrow\mathcal{M}}
\mathbf{Q}_{k,i}^{I\rightarrow\mathcal{M}},
\qquad
\mathbf{g}_{k,\mathrm{max}}^{I\rightarrow\mathcal{M}}
=
\max_{i=1,\ldots,V_r}
\mathbf{Q}_{k,i}^{I\rightarrow\mathcal{M}}.
\label{eq:app_rde_pooling}
\end{equation}
The confidence-weighted mean captures the dominant discrepancy trend across
reliable ray correspondences, while max pooling preserves the most salient
local mismatches that may indicate critical residual pose corrections. Together,
they summarize both global consistency and local evidence in the ray
discrepancy pattern, providing a compact condition for residual pose prediction.
We perform the same aggregation for the floorplan-to-image
direction and concatenate the two directional pooled features:
\begin{equation}
\begin{aligned}
\mathbf{z}_{k}
&=
\Phi_{\mathrm{RDE}}
\left(
\mathbf{F}^{I},
\mathbf{F}_{k}^{\mathcal{M}},
\hat{\mathbf{r}}^{I},
\mathbf{r}_{k}^{\mathcal{M}}
\right)
\\
&=
\bigl[
\mathbf{g}_{k,\mathrm{mean}}^{I\rightarrow\mathcal{M}},
\mathbf{g}_{k,\mathrm{max}}^{I\rightarrow\mathcal{M}},
\mathbf{g}_{k,\mathrm{mean}}^{\mathcal{M}\rightarrow I},
\mathbf{g}_{k,\mathrm{max}}^{\mathcal{M}\rightarrow I}
\bigr].
\end{aligned}
\label{eq:app_rde_global_condition}
\end{equation}
The resulting feature $\mathbf{z}_{k}$ compactly summarizes the bidirectional
ray discrepancy under the $k$-th coarse anchor and serves as the geometric
condition for the diffusion denoiser.

Overall, RDE is designed to extract residual pose cues from ray-level
discrepancies. The soft correspondence mechanism improves robustness to
imperfect coarse anchors and noisy ray observations, bidirectional matching
captures complementary evidence from both image-to-floorplan and
floorplan-to-image perspectives, and confidence-aware aggregation guides the
denoiser to focus on reliable yet informative discrepancies for residual pose
prediction.

\subsection{Real-World Dataset Details}
\label{supp:real_world_dataset_details}

Following \fthreeloc, we adapt LaMAR~\cite{sarlin2022lamar} to our real-world floorplan localization setting by constructing a subset with paired RGB observations, ground-truth poses, and a corresponding 2D floorplan. 
%The original LaMAR dataset contains sequences captured by both iOS and HoloLens devices. 
We use the iOS sequences from the HGE building. Since some trajectories extend beyond the building interior, we filter the data using the provided ground-truth poses and the spatial extent of the HGE building, retaining only indoor observations that fall within the HGE floorplan coverage, where floorplan-based localization is well-defined.

We obtain the complete HGE building floorplan from the official source and use it as the localization map. To align the original LaMAR annotations with our floorplan localization setting, we convert the provided $\mathrm{SE}(3)$ poses from the global world coordinate system into $\mathrm{SE}(2)$ poses in the floorplan coordinate system. Specifically, we manually select four salient landmarks that can be reliably identified in both the RGB observations and the floorplan. For each landmark, we record its 2D position in the LaMAR world coordinate system and its corresponding pixel coordinate on the floorplan. These four point correspondences provide eight constraints, which are sufficient to estimate the six parameters of a 2D affine transformation in a least-squares~\cite{bjorck1990least} manner. We then apply the estimated affine transformation to all ground-truth trajectory positions. For orientation, we extract the yaw heading from the original quaternion and transform it consistently into the floorplan coordinate system using the linear part of the estimated affine transformation. This yields the final $\mathrm{SE}(2)$ pose annotations on the floorplan.

To better match the floorplan geometry to the visual observations, we manually remove non-geometric annotations such as room numbers and stair labels. We also remove doors at traversable connections along the collected trajectories while
preserving other doors as solid walls, since they are mostly closed during LaMAR data collection. Landmark selection, floorplan alignment, and preprocessing use no validation or test trajectories and are performed solely to adapt LaMAR to the floorplan-localization setting, rather than to tune the
localization method. After preprocessing, we split the retained data into training, validation, and test sets. Quantitative evaluations of all methods on this real-world setting are reported in Sec.~\ref{sec:real_quan_exp}.

\section{Additional Qualitative Results}
\label{supp:qualitative_results}
\subsection{Diffusion-Aided Pose Refinement Results}
\label{supp:pose_refinement_qualitative}

\begin{figure}[t]
    \centering
    \includegraphics[width=0.98\linewidth]{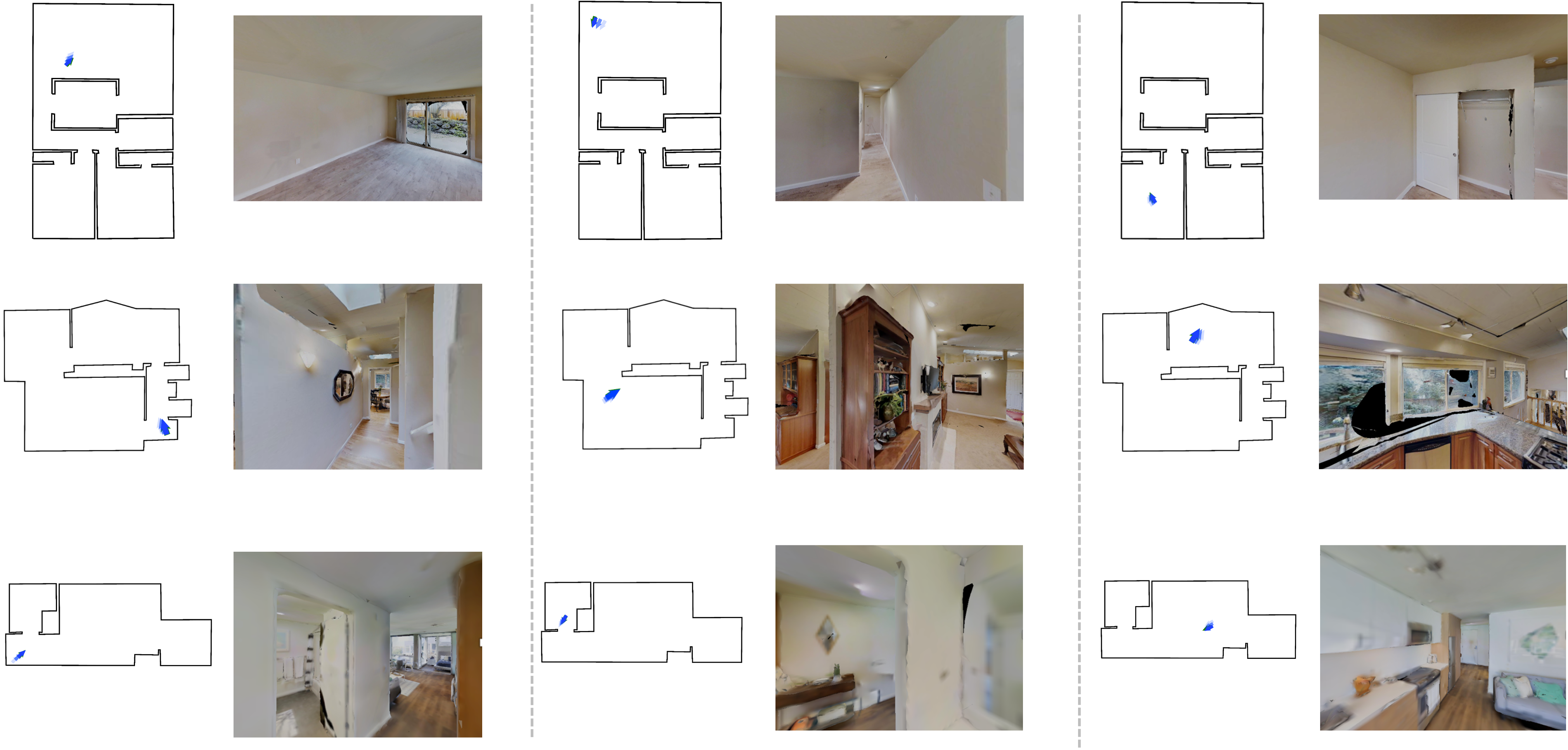}
    \caption{
    \textbf{Qualitative results of diffusion-aided pose refinement.}
Starting from coarse pose anchors retrieved by online floorplan sampling, our 
diffusion-aided pose refinement module progressively predicts residual pose 
corrections guided by paired image-side and floorplan-side ray geometry. The 
color transition from light blue to dark blue indicates the pose 
recovery process, while green denotes the ground truth. Coarse anchors from 
different initial locations are refined toward poses closer to the ground truth, 
validating the effectiveness of diffusion-aided pose refinement.
    }
    \label{fig:pose_refinement_qualitative}
    \vspace{-1.0em}
\end{figure}

Fig.~\ref{fig:pose_refinement_qualitative} presents additional qualitative results of our diffusion-aided pose refinement module. The results show that FreeLoc consistently refines coarse pose anchors toward the ground-truth poses across different scenes and locations. Even when the initial coarse samples are located at varying distances from the target pose, the refinement module effectively corrects their residual translation and orientation errors, producing refined hypotheses that are closer to the ground truth. These qualitative results demonstrate that the proposed module can move beyond discretized coarse samples and perform accurate continuous pose optimization.

\subsection{Single-Frame Localization Results}
\label{supp:single_frame_qualitative}

\begin{figure}[t]
    \centering
    \includegraphics[width=0.98\linewidth]{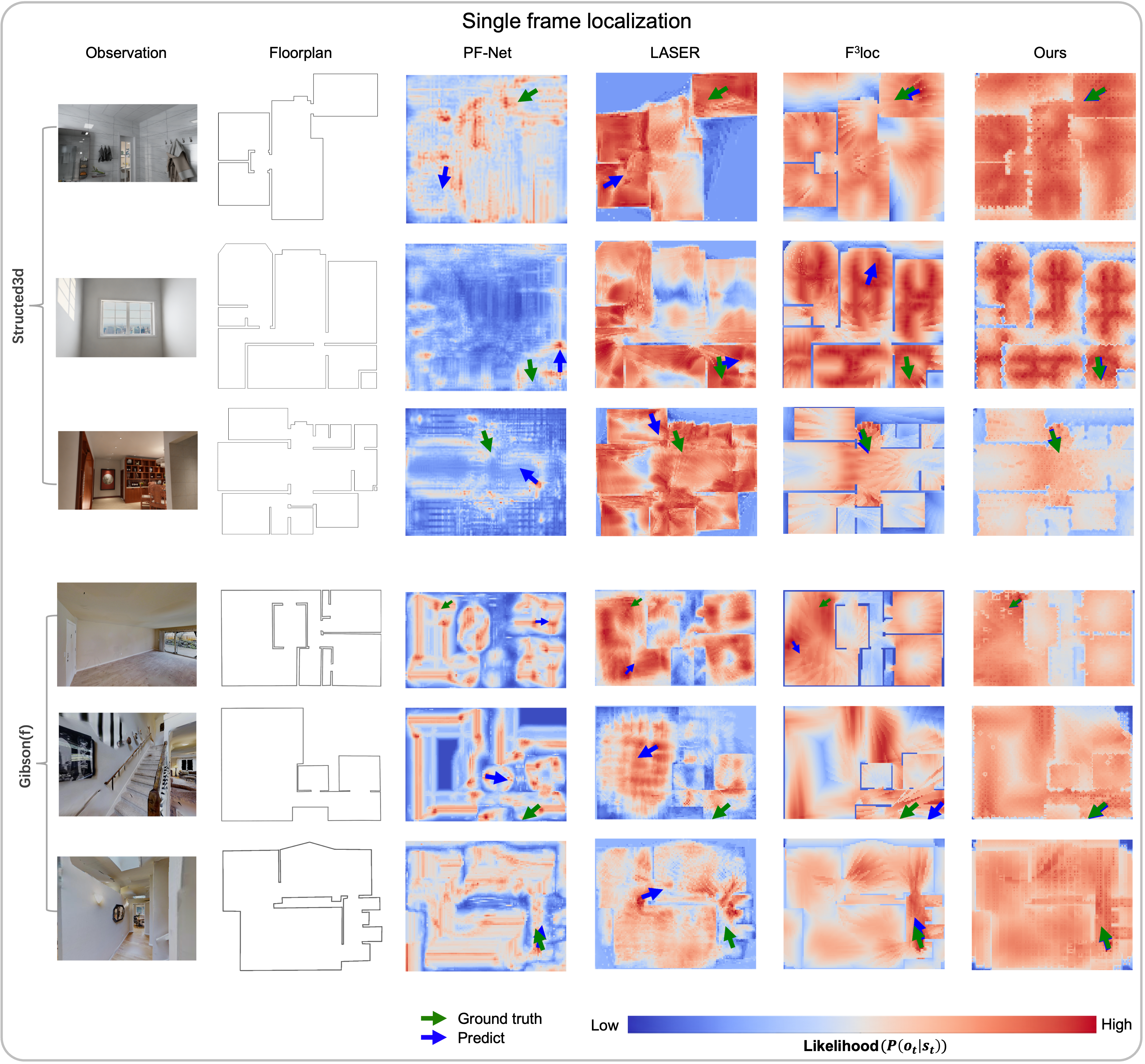}
    \caption{
   \textbf{Qualitative comparison of single-frame localization results.}
We compare \ours with representative baselines on Gibson and Structured3D.
Compared with PF-Net, LASER, and F$^3$Loc, \ours produces pose estimates that
are better aligned with the ground truth, benefiting from accurate online coarse
pose retrieval and diffusion-aided pose refinement.
    }
    \label{fig:single_frame_qualitative}
    \vspace{-1.0em}
\end{figure}

 Fig.~\ref{fig:single_frame_qualitative} presents qualitative single-frame localization results on Gibson and Structured3D. The results show that FreeLoc produces more accurate pose estimates than PF-Net~\cite{pf-net}, LASER~\cite{semantic-based:min2022laser}, and \fthreeloc~\cite{chen2024f3loc} across different scenes. Without relying on a scene-specific offline ray database, FreeLoc achieves better single-frame localization than \fthreeloc{} by retrieving reliable coarse pose anchors online and further refining them in continuous pose space through diffusion-aided pose refinement.

\subsection{Sequential Localization Results}
\label{supp:sequential_qualitative}

\begin{figure}[t]
    \centering
    \includegraphics[width=0.98\linewidth]{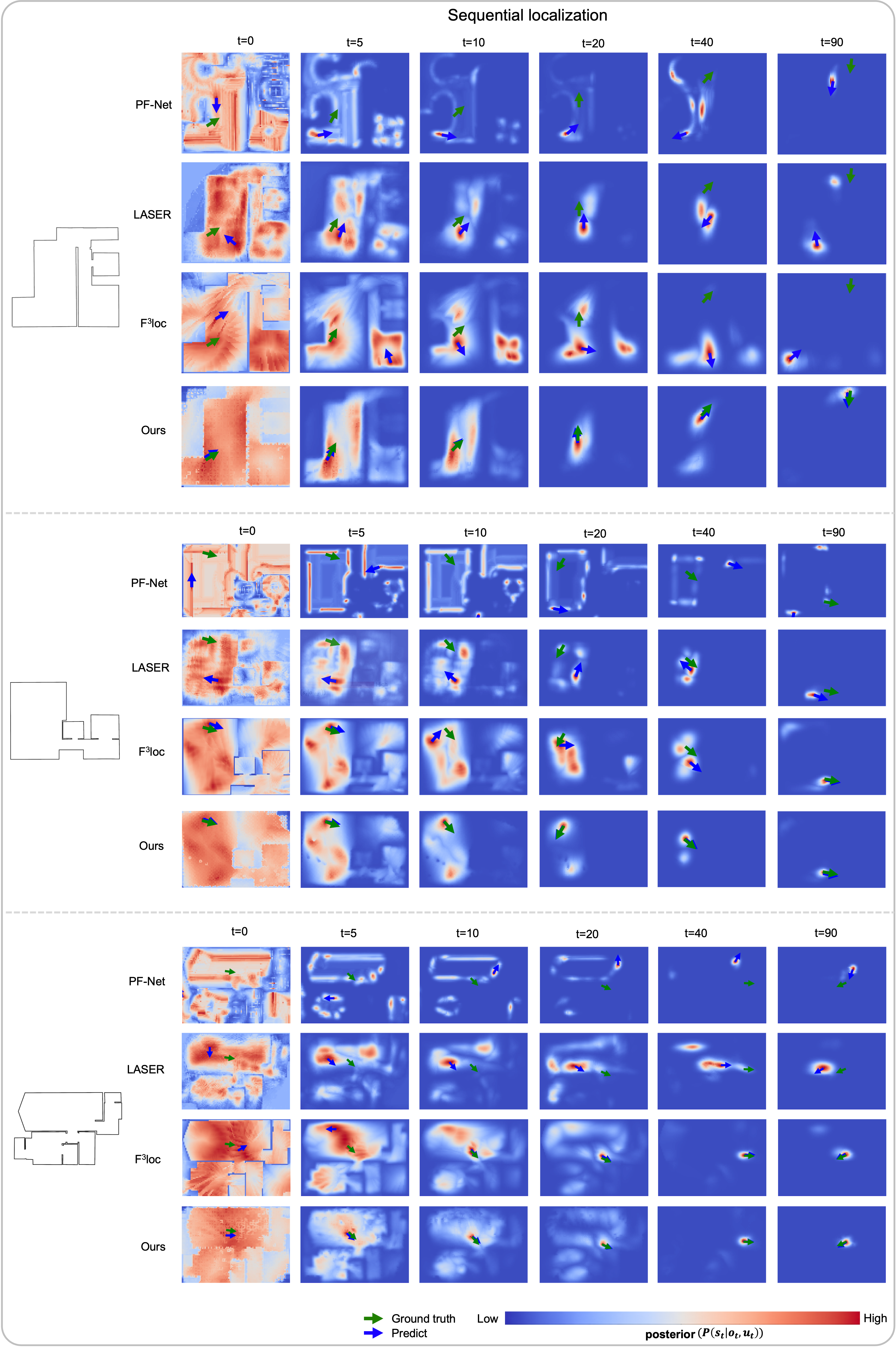}
    \caption{
\textbf{Qualitative comparison of sequential localization results.}
We compare \ours with representative baselines on the Gibson(t) sequential 
localization dataset. The results show the temporal evolution of posterior 
distributions. Benefiting from more reliable single-frame localization evidence, 
\ours concentrates the posterior around the correct region more quickly and 
maintains more accurate localization over time.
    }
    \label{fig:sequential_qualitative}
    \vspace{-1.0em}
\end{figure}

Fig.~\ref{fig:sequential_qualitative} presents qualitative sequential localization results across different scenes. The results show that FreeLoc converges faster and maintains more accurate localization than the baselines. By constructing observation likelihoods online from reliable coarse anchors and diffusion-refined candidates, FreeLoc provides more accurate frame-level evidence for temporal filtering, allowing the posterior distribution to concentrate around the correct region more quickly and remain stable over time. Unlike \fthreeloc{}, which relies on a scene-specific offline ray database, our online likelihood construction improves deployment flexibility in new environments.

\section{Additional  Quantitative Results}
\label{supp:quantitative_results}
\subsection{Real-World Results}
\label{sec:real_quan_exp}

Table~\ref{tab:real_world_quantitative} presents the quantitative results on the real-world HGE building dataset. The HGE building is a challenging floorplan localization setting, covering a large spatial extent of approximately $130\,\mathrm{m}\times130\,\mathrm{m}$ with complex indoor layouts and repetitive structural patterns. As a result, all methods achieve relatively low single-frame recall in this real-world scenario. Moreover, both \fthreeloc{} and our method rely on image-side floorplan-depth prediction. In such a large-scale real-world environment, ray depths can span tens of meters in many regions, which substantially increases depth-estimation uncertainty and limits the overall reliability of single-frame localization. Despite this difficulty, depth-based methods can still produce relatively concentrated single-frame likelihoods in regions with moderate ray depths and sufficient geometric constraints. These likelihoods provide useful frame-level evidence for temporal filtering, enabling sequential localization to reach a reliable level of accuracy over time.

For single-frame localization, our method achieves the best performance under the stricter thresholds, especially at $0.1~\mathrm{m}$ and $0.5~\mathrm{m}$. This demonstrates the benefit of diffusion-aided pose refinement, which optimizes coarse anchors in the continuous pose space and produces more accurate pose hypotheses around plausible locations.

For sequential localization, the concentrated single-frame likelihoods are accumulated through temporal filtering, substantially improving localization stability. Our method achieves the same $\mathrm{SR}@1\mathrm{m}$ as \fthreeloc{} while reducing $\mathrm{RMSE}_{\mathrm{all}}$ from $15.1~\mathrm{m}$ to $14.5~\mathrm{m}$. The $\mathrm{RMSE}_{\mathrm{succ}}$ of our method is slightly higher than \fthreeloc~(0.4 m vs. 0.2 m). This small gap mainly comes from reconstructing dense observation likelihoods 
by interpolating a compact set of online-retrieved and refined pose-score 
candidates, whereas \fthreeloc{} evaluates likelihoods over a dense offline ray 
database. Such interpolation may slightly reduce local precision for already 
successful trajectories, but it enables database-free online localization while 
maintaining competitive sequential accuracy in real-world deployment.
\begin{table}[t]
\centering
\caption{Real-world quantitative localization results on the HGE building.
-- indicates that no trajectory was successfully localized.
}
\label{tab:real_world_quantitative}
\footnotesize
\setlength{\tabcolsep}{3.0pt}
\renewcommand{\arraystretch}{1.05}

\begin{adjustbox}{max width=\linewidth}
\begin{tabular}{lccccccc}
\toprule
\multirow{2}{*}{\textbf{Method}}
& \multicolumn{4}{c}{\textbf{Single-frame Localization}}
& \multicolumn{3}{c}{\textbf{Sequential Localization}} \\
\cmidrule(lr){2-5} \cmidrule(lr){6-8}
& $@0.1\mathrm{m}\uparrow$
& $@0.5\mathrm{m}\uparrow$
& $@1\mathrm{m}\uparrow$
& $@1\mathrm{m}30^\circ\uparrow$
& $\mathrm{SR}@1\mathrm{m}$ (\%) $\uparrow$
& $\mathrm{RMSE}_{\mathrm{succ}}$ (m) $\downarrow$
& $\mathrm{RMSE}_{\mathrm{all}}$ (m) $\downarrow$ \\
\midrule

PF-Net
& 0
& 0
& 0
& 0
& 0
& --
& 76.9 \\

LASER
& 0
& 0.3
& 0.7
& 0.7
& 0
& --
& 41.4 \\

F$^3$Loc
& 0
& 0
& \textbf{1.7}
& \textbf{1.7}
& \textbf{66.7}
& \textbf{0.2}
& 15.1 \\

Ours
& \textbf{0.3}
& \textbf{1.3}
& \textbf{1.7}
& \textbf{1.7}
& \textbf{66.7}
& 0.4
& \textbf{14.5} \\
\bottomrule
\end{tabular}
\end{adjustbox}

\vspace{-1.0em}
\end{table}

\subsection{Ablation Study of Diffusion-Aided Pose Refinement}
\label{supp:ablation}

We first compare our diffusion-aided pose refinement with ICP optimization in
the main paper. Since residual pose refinement can naturally be formulated as
a geometric registration problem, ICP serves as a representative deterministic
alternative. However, ICP relies on reliable geometric correspondences and
local alignment, whereas our ray observations are sparse and imperfect due to
coarse pose errors and image-side ray prediction noise, limiting the
effectiveness of deterministic alignment. This comparison demonstrates the
benefit of learned refinement over conventional geometric optimization, but
does not isolate whether the gain originates from the learned ray-discrepancy
representation or the diffusion formulation itself. We therefore further
introduce a simpler learned direct-regression baseline to specifically examine
the contribution of diffusion-based refinement.

For a controlled comparison, both learned variants use the same RDE to
construct the global ray-discrepancy feature, with matched network capacity and
identical training settings. The only difference lies in the refinement
formulation: the direct-regression variant replaces iterative diffusion
denoising with a one-shot regression head that directly predicts the relative
pose correction.

As shown in Table~\ref{tab:ablation_regression}, diffusion-aided refinement
consistently outperforms direct regression across all reported metrics,
including a $2.8$ percentage-point gain in SR@0.2m and a $25\%$ reduction in
RMSE$_{\rm all}$. Since the feature representation, model capacity, and
training configuration are controlled, these improvements directly isolate
the benefit of the diffusion-based refinement formulation. At the same time,
the competitive performance of direct regression indicates that the RDE
provides informative geometric conditioning.

Compared with one-shot regression, iterative denoising is better suited to
residual pose optimization by progressively refining pose corrections toward
geometrically consistent solutions. Moreover, noisy and locally ambiguous
ray-level discrepancies make a single deterministic prediction less flexible
in capturing uncertainty in the residual correction. By modeling the residual
correction as a conditional distribution and progressively denoising it under
geometric guidance, diffusion-aided refinement produces more accurate pose
hypotheses. 
Together, the ICP and direct-regression comparisons show that the
performance gains arise not only from learned geometric representations, but
also from the diffusion-based refinement formulation itself.
% \begin{table}[!t]
% \centering
% \caption{Ablation study of diffusion-aided pose refinement against direct regression.}
% \label{tab:ablation_regression}
% \footnotesize
% \setlength{\tabcolsep}{4pt}
% \renewcommand{\arraystretch}{1.05}
% \begin{tabular}{lcccc}
% \toprule
% \multirow{2}{*}{\textbf{Method}}
% & \multicolumn{4}{c}{\textbf{Recall (\%)}} \\
% \cmidrule(lr){2-5}
% & $@0.1\mathrm{m}\uparrow$
% & $@0.5\mathrm{m}\uparrow$
% & $@1\mathrm{m}\uparrow$
% & $@1\mathrm{m}30^\circ\uparrow$ \\
% \midrule
% Direct Regression
% & 8.9
% & 36.4
% & 42.0
% & 40.3 \\

% Diffusion-Aided Pose Refinement
% & \textbf{9.5}
% & \textbf{37.0}
% & \textbf{43.0}
% & \textbf{40.9} \\
% \bottomrule
% \end{tabular}
% \vspace{-0.5em}
% \end{table}

\begin{table}[t]
\centering
\caption{Ablation study of diffusion-aided pose refinement against direct regression on Gibson(t).}
\label{tab:ablation_regression}

\vspace{-0.5em}

\scriptsize
\setlength{\tabcolsep}{2.8pt}
\renewcommand{\arraystretch}{1.05}

\resizebox{0.8\linewidth}{!}{%
\begin{tabular}{lcccc}
\toprule
\textbf{Refinement}
& \textbf{SR@1m (\%)} $\uparrow$
& \textbf{SR@0.5m (\%)} $\uparrow$
& \textbf{SR@0.2m (\%)} $\uparrow$
& \textbf{RMSE$_{\rm all}$ (m)} $\downarrow$ \\
\midrule

Direct Regression
& 97.3
& 97.3
& 48.6
& 0.20 \\

Diffusion-Aided Pose Refinement
& \textbf{100.0}
& \textbf{100.0}
& \textbf{51.4}
& \textbf{0.15} \\

\bottomrule
\end{tabular}%
}

\vspace{-2.0em}
\end{table}

\subsection{Compatibility with Stronger Depth Estimation}
\label{supp:unloc_exp}

We further compare with UnLoc~\cite{wuest2025unloc}, which extends F$^3$Loc
with stronger foundation-model-based monocular depth estimation using Depth
Anything V2~\cite{depth_anything_v2}. Since depth quality directly affects the image-side geometric
observations, we use the same UnLoc depth predictions for FreeLoc to enable a
controlled comparison of the localization pipelines. Specifically, we replace
our default depth predictions with those from UnLoc at test time while keeping
the FreeLoc pipeline and diffusion refinement model unchanged, without
retraining.

As shown in Table~\ref{tab:unloc_comparison}, FreeLoc consistently outperforms
UnLoc across all single-frame metrics on both Gibson(f) and Structured3D, with
gains of $0.6$--$3.1$ percentage points and $0.4$--$1.8$ percentage points,
respectively. In sequential localization, FreeLoc further improves the success
rate from $97.3\%$ to $100.0\%$ and reduces RMSE$_{\rm all}$ from
$0.28\,\mathrm{m}$ to $0.15\,\mathrm{m}$. Since both methods use the same depth
predictions, these results demonstrate that FreeLoc maintains superior
localization accuracy under stronger depth estimation. Moreover, directly
adopting UnLoc depth without retraining demonstrates the plug-and-play
compatibility of our pipeline with stronger monocular depth estimators.

\begin{table*}[h]
\centering
\caption{
Controlled comparison with UnLoc under matched depth predictions.
FreeLoc uses the same UnLoc depth predictions at test time without retraining
the diffusion refinement model. Bold denotes the better result.
}
\label{tab:unloc_comparison}

\vspace{-0.5em}
\small
\renewcommand{\arraystretch}{1.06}

% ----------------------------------------------------------------
% (a) Single-frame localization
% ----------------------------------------------------------------
\textbf{(a) Single-frame localization}\par
\vspace{0.3em}

\setlength{\tabcolsep}{4.0pt}
\resizebox{0.82\textwidth}{!}{%
\begin{tabular}{lcccccccc}
\toprule
\multirow{2}{*}{\textbf{Method}}
& \multicolumn{4}{c}{\textbf{Gibson(f)}}
& \multicolumn{4}{c}{\textbf{Structured3D}} \\
\cmidrule(lr){2-5}
\cmidrule(lr){6-9}

& $@0.1\mathrm{m}\uparrow$
& $@0.5\mathrm{m}\uparrow$
& $@1\mathrm{m}\uparrow$
& $@1\mathrm{m}30^\circ\uparrow$
& $@0.1\mathrm{m}\uparrow$
& $@0.5\mathrm{m}\uparrow$
& $@1\mathrm{m}\uparrow$
& $@1\mathrm{m}30^\circ\uparrow$ \\
\midrule

UnLoc
& 13.5
& 46.1
& 49.6
& 47.8
& 3.6
& 27.7
& 33.6
& 32.6 \\

\textbf{Ours}
& \textbf{16.6}
& \textbf{46.9}
& \textbf{50.2}
& \textbf{48.7}
& \textbf{4.9}
& \textbf{29.5}
& \textbf{34.3}
& \textbf{33.0} \\

\bottomrule
\end{tabular}%
}

\vspace{0.8em}

% ----------------------------------------------------------------
% (b) Sequential localization
% ----------------------------------------------------------------
\textbf{(b) Sequential localization on Gibson(t)}\par
\vspace{0.3em}

\setlength{\tabcolsep}{5.0pt}
\resizebox{0.50\textwidth}{!}{%
\begin{tabular}{lccc}
\toprule
\textbf{Method}
& $\mathrm{SR}@1\mathrm{m}$ (\%) $\uparrow$
& $\mathrm{RMSE}_{\mathrm{succ}}$ (m) $\downarrow$
& $\mathrm{RMSE}_{\mathrm{all}}$ (m) $\downarrow$ \\
\midrule

UnLoc
& 97.3
& 0.16
& 0.28 \\

\textbf{Ours}
& \textbf{100.0}
& \textbf{0.15}
& \textbf{0.15} \\

\bottomrule
\end{tabular}%
}

\end{table*}

\end{document}